\documentclass[lettersize,journal]{IEEEtran}
\usepackage{textcomp}
\usepackage{verbatim}
\usepackage{url}            
\usepackage{amsfonts}       
\usepackage{microtype}      
\usepackage{amssymb}
\usepackage{amsmath}        

\usepackage{cite}

\usepackage[caption=false,font=normalsize,labelfont=sf,textfont=sf]{subfig}
\usepackage{graphicx}
\usepackage{stfloats} 

\usepackage[table]{xcolor}
\usepackage{array}
\usepackage{multirow}
\usepackage{arydshln}
\usepackage{colortbl}

\newtheorem{definition}{Definition}

\usepackage[ruled,linesnumbered]{algorithm2e}

\usepackage{orcidlink}
\usepackage{cleveref} 

\begin{document}

\title{Adversarial Training for Graph Neural Networks via Graph Subspace Energy Optimization}

%

\author{Ganlin Liu\,\orcidlink{0009-0002-0680-1055}, \and Ziling Liang\,\orcidlink{0009-0005-3956-0421}, \and Xiaowei Huang\,\orcidlink{0000-0001-6267-0366}, \and Xinping Yi\,\orcidlink{0000-0001-5163-2364}, \textit{Member, IEEE}, \and Shi Jin\,\orcidlink{0000-0003-0271-6021}, \textit{Fellow, IEEE}
%

\thanks{G. Liu and X. Huang are with the Department of Computer Science, The University of Liverpool, Liverpool, UK. Email: \{ganlin.liu,~xiaowei.huang\}@liverpool.ac.uk}
\thanks{Z. Liang, X. Yi, and S. Jin are with the National Mobile Communications Research Laboratory, Southeast University, Nanjing 210096, China. Email: \{zlliang, xyi, jinshi\}@seu.edu.cn}
}

\maketitle
\begin{abstract}
Despite impressive capability in learning over graph-structured data, graph neural networks (GNN) suffer from adversarial topology perturbation in both training and inference phases. While adversarial training has demonstrated remarkable effectiveness in image classification tasks, its suitability for GNN models has been doubted until a recent advance that shifts the focus from \emph{transductive} to \emph{inductive} learning. Still, GNN robustness in the inductive setting is under-explored, and it calls for deeper understanding of GNN adversarial training. 
To this end, we propose a new concept of graph subspace energy (GSE)---a generalization of graph energy that measures graph stability---of the adjacency matrix, as an indicator of GNN robustness against topology perturbations.
To further demonstrate the effectiveness of such concept, we propose an adversarial training method with the perturbed graphs generated by maximizing the GSE regularization term, referred to as AT-GSE.
To deal with the local and global topology perturbations raised respectively by LRBCD and PRBCD, we employ randomized SVD (RndSVD) and Nystr\"om low-rank approximation to favor the different aspects of the GSE terms.
An extensive set of experiments shows that AT-GSE outperforms consistently the state-of-the-art GNN adversarial training methods over different homophily and heterophily datasets in terms of adversarial accuracy, whilst more surprisingly achieving a superior clean accuracy on non-perturbed graphs.
\end{abstract}


\section{Introduction}
\IEEEPARstart{R}{cent} years have witnessed the remarkable effectiveness of graph neural networks (GNNs) in processing complex graph-structured data, such as social networks \cite{wu2020graph}, physical systems \cite{sanchez2018graph}, knowledge graphs \cite{hamaguchi2017knowledge}, and many others (e.g., \cite{khalil2017learning}). 
Graph convolution network (GCN) \cite{GCN} is one of the pioneering GNN models that leverage only local information aggregation, yielding superior accuracy performance in node classification, graph classification, and link prediction tasks. 
Albeit promising in many tasks, it is noticed that GNN models are also vulnerable to adversarial attacks, such as label poisoning attacks (e.g., LafAK \cite{LafAK} and MGattack \cite{MG}) and graph topology poisoning and evasion attacks (e.g., Metattack \cite{Metattack}, Nettack \cite{Nettack}, PRBCD \cite{PRBCD} and LRBCD \cite{GD-AT}). Such adversarial attacks could either mislead the training process to end up with a toxic GNN model (cf. poisoning attacks) or cause well-trained GNN models degraded performance in certain situations (cf. evasion attacks). 

To defend against these adversarial attacks, there are an increasing number of studies that attempt to protect the GNN model architectures or training process. 
For the graph poisoning attacks (e.g., Metattack and Nettack), a number of robust GNN models, such as GCN-Jaccard \cite{GCN-Jaccard}, GCN-SVD \cite{GCN-SVD}, SimP-GCN \cite{SimP-GCN} and Pro-GNN \cite{Pro-GNN}, have been proposed to pre-process the perturbed training graphs via graph properties, such as feature similarity, sparsity and low-rankness of adjacency matrix to improve the adversarial robustness. For graph evasion attacks (e.g., PRBCD and LRBCD), adversarial training (AT) has become a common method to improve GNN's resistance against adversarial attacks without sacrificing classification accuracy as much as possible. Nevertheless, most adversarial training methods, such as Xu et al. \cite{2xu2019topology}, Xu et al.\cite{3xu2020towards}, Deng et al. \cite{34deng2023batch}, Feng et al. \cite{35feng2019graph}, Jin and Zhang\cite{36jin2019latent}, Chen et al. \cite{37chen2020smoothing}, Li et al. \cite{38li2022spectral}, and Guo et al. \cite{39guo2022learning}, are based on the transductive learning setting, which is impractical in real-world scenarios because it is unlikely the test set is informed in advance. A recent study \cite{GD-AT} reveals both the theoretical and empirical limitations of adversarial training in the transductive learning setting, calling for rethinking adversarial training in a fully inductive learning setting, where test nodes are unavailable during training. This is the focus of the current work.
\begin{figure}[!ht]
\renewcommand\thesubfigure{\roman{subfigure}}
    \centering
    \includegraphics[width=0.7\linewidth]{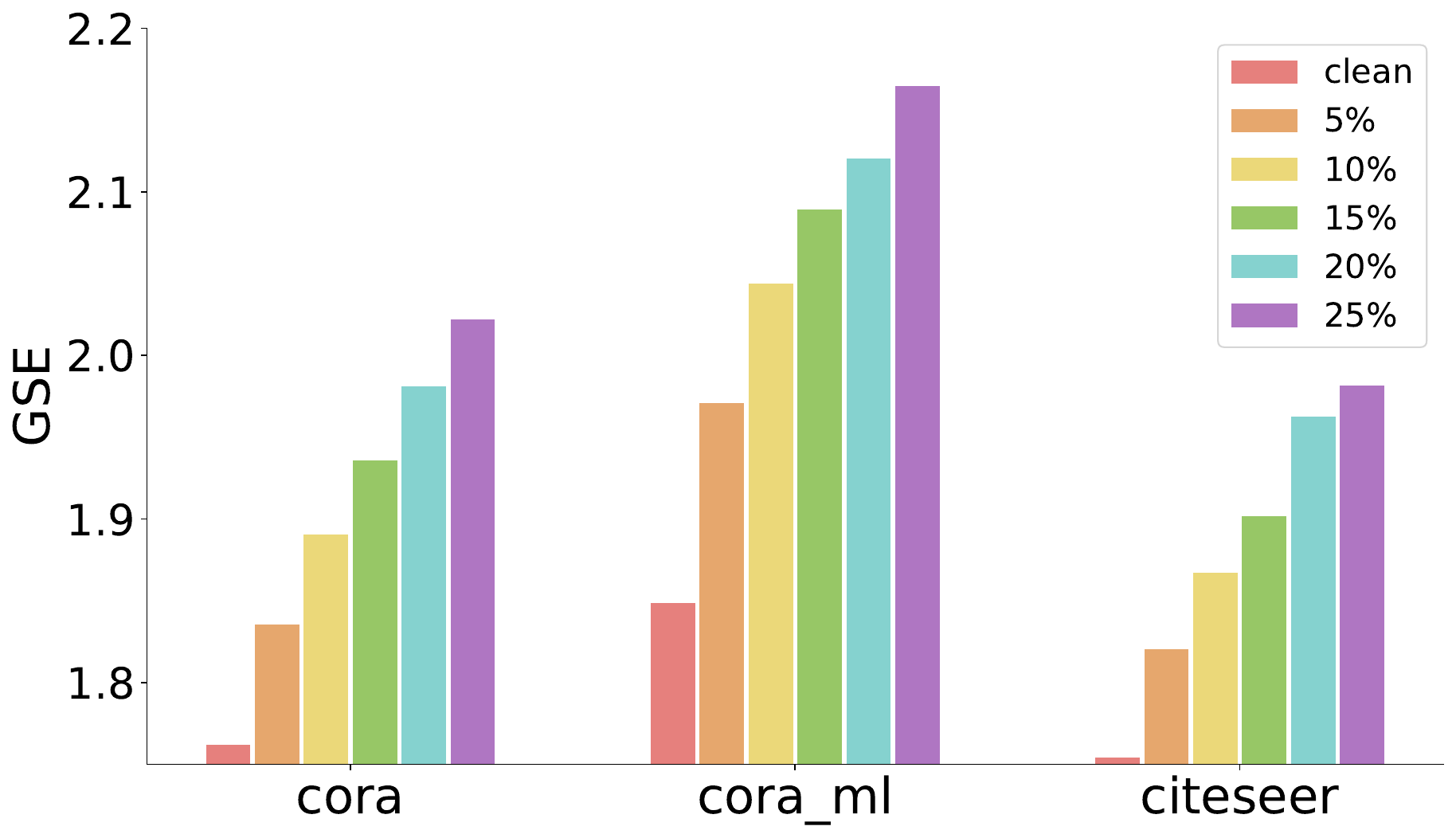}
    \caption{Normalized graph subspace energy (GSE) of different datasets under adversarial topology attacks with different attack ratios.} 
    \label{fig:GSE_bar}
\end{figure}

In this paper, we take a look at graph topology evasion attacks from a graph energy perspective, and investigate adversarial training for GNN models in the fully inductive setting. In particular, given different graph datasets (e.g., Cora, Cora\_ML, Citeseer) under Metattack, we observe, as shown in \Cref{fig:GSE_bar}, that the normalized graph energy exhibits certain property: the stronger the attack, the larger the graph subspace energy (GSE), a generalized concept of graph energy whose definition is in \Cref{def:GSE}. 
This suggests that graph energy could serve as a measurable indicator of graph topology attacks, through which we could be able to control the levels of graph topology attacks (such as Metattack). 
When it comes to adversarial training for graph neural networks, it would be possible to produce worst-case adversarial topology perturbations by maximizing the corresponding graph energy, instead of resorting to gradient-based attacks as for the image classification tasks. Throughout the paper, we aim to show how graph energy can be employed in adversarial training and to what extent it can improve adversarial robustness of state-of-the-art GNN models. 
Specifically, our main \textit{contributions} are summarized as follows:

\begin{itemize}
    \item We propose a new concept of graph subspace energy (GSE), which can serve as a measurable indicator of GNN robustness against adversarial topology perturbation. The GSE is a generalization of graph energy, specifying graph energy contained in a subspace spanned by a set of singular vectors corresponding to a certain range of singular value distribution of the adjacency matrix. It is consistently observed adversarial topology attacks decrease node classification accuracy with the GSE increased simultaneously.
    \item Towards inductive learning of GNNs, we propose a novel adversarial training (AT) method with a minimax optimization formalism, referred to as AT-GSE, where the inner maximization aims to produce the worst-case graph perturbation with respect to the maximal GSE, and the outer minimization finds the model parameters over perturbed graphs via stochastic gradient descent (SGD) training. 
    Notably, AT-GSE can be employed flexibly in various GNN models to improve adversarial robustness.
    \item To enhance AT-GSE against the local (e.g., LRBCD) and global (e.g., PRBCD) topology perturbations, we employ the randomized SVD (RndSVD) and Nystr\"om low-rank approximation methods for the computations of the GSE terms, which place local and global emphases on graph spectrum, respectively. The former transforms the matrix into a smaller matrix that can be processed faster through randomization and QR decomposition, while the latter uses sampling and pseudo-inverse to obtain a low-rank approximation. They both increase efficiency and save storage space without sacrificing performance.
\end{itemize}
Through an extensive set of experiments, AT-GSE can improve the adversarial robustness of various datasets over the state-of-the-art GNN defending methods with a large margin. At the same time, GSE as a regularizer can consistently improve model clean accuracy across different GNN models. Also, RndSVD and Nystr\"om methods have efficient and excellent performance in defending against local and global attacks respectively.
An extensive set of experiments on 7 datasets containing both homophilic and heterophilic graphs demonstrate that our proposed AT-GSE has superior robustness, generalization, and scalability performance.

The rest of the paper is organized as follows. In \Cref{sec:background}, we introduce the closely related work, including both adversarial attacks and defenses in the literature. 
\Cref{sec:AT-GSE} formally defines graph subspace energy and describes our proposed AT-GSE in detail. To reduce complexity, it also presents two methods: RndSVD and Nystr\"om.
In \Cref{sec:experiments}, experiments on 7 datasets are presented to compare our method with state-of-the-art ones with respect to both clean and adversarial accuracy. 
Finally, we conclude the paper in \Cref{sec:conclusion}.

\section{Background}
\label{sec:background}
Although powerful in node classification tasks, graph neural networks (GNNs) are vulnerable to graph topology adversarial attacks (\Cref{sub:Adversarail_Attacks}), which cause model performance degradation. To defend against such attacks, there exist various methods that resist adversarial attacks and improve model robustness (cf. adversarial defenses in \Cref{sub:Adversarail_Defenses}).

\subsection{Adversarial Attacks}
\label{sub:Adversarail_Attacks}
With respect to different targets of graph data, adversarial attacks on graphs can be categorized into label attacks, node attribute attacks and graph topology attacks. 
These attacks aim to exploit vulnerabilities of the GNN models to what extent the modified label/attribute/topology data could result in misclassification or incorrect predictions.

The goal of \textbf{label attacks}, such as LafAK \cite{LafAK} and MGattack \cite{MG}, is mainly to mislead the model classification as much as possible by flipping a few target labels. 
In contrast, adversarial topology attacks pay more attention to modifying the graph structure. More specifically, adversarial topology attacks can be divided into poisoning and evasion attacks according to different stages when attacks happen. \textbf{Poisoning attacks} occur before training the model. There are two representative poisoning attacks: Metattack \cite{Metattack} treats the graph structure as a hyper-parameter to strategically modify the graph topology; Nettack \cite{Nettack} iteratively identifies which edge modifications will cause the largest degradation in model performance. 
It is worth noting that some poisoning attacks are also applicable to the inference phase as evasion attacks, where the latter is our focus.

Our focus in this paper is placed on \textbf{evasion attacks}, where the topology attack  happens on well-trained GNN models in the inference phase. There are only a few graph evasion attacks in the literature, such as projected randomized block coordinate descent (PRBCD) \cite{PRBCD} and locally constrained randomized block coordinate descent (LRBCD) \cite{GD-AT}. The PRBCD attack is a gradient-based attack framework building upon a variant of the classic block coordinate descent (BCD) algorithm. In PRBCD, optimization variables are divided into blocks, with blocks of variables randomly selected to update on each iteration --- such randomization helps convergence --- followed by a projection step to ensure that the updated variables satisfy any constraints or bounds. As an extension of PRBCD, LRBCD adds local constraints that apply to specific subsets of variables rather than just the global objective function, yielding efficient attacks containing local constraints.

\subsection{Adversarial Defenses}
\label{sub:Adversarail_Defenses}
To protect graph neural networks from the adversarial attacks, different \textbf{robust models} have been proposed. GCN-Jaccard \cite{GCN-Jaccard}, SimP-GCN \cite{SimP-GCN}, and GCN-SVD \cite{GCN-SVD} preprocess the adjacency matrix of the graph and use similarity or low-rank approximation to filter noise. Pro-GNN \cite{Pro-GNN} adds some regularizers to the loss function to protect the low-rankness and sparsity of the graph properties. Robust-GCN \cite{Robust-GCN} uses Gaussian distribution as the hidden representation in the convolutional layer to absorb the effects of adversarial attacks. This type of robust models resists poisoning attacks. 

For evasion attacks, a preferred method is \textbf{adversarial training (AT)}. The main goal of AT is to enhance the robustness and generalization of a model by exposing it to adversarial examples during training. Many studies verify that it is effective in improving the classification accuracy on clean graphs and the robustness of machine learning against various adversarial attacks. At present, AT has many variants based on adversarial examples generated by different attacks, such as Xu et al. \cite{2xu2019topology}, Xu et al.\cite{3xu2020towards}, Deng et al. \cite{34deng2023batch}, Feng et al. \cite{35feng2019graph}, Jin and Zhang\cite{36jin2019latent}, Chen et al. \cite{37chen2020smoothing}, Li et al. \cite{38li2022spectral}, and Guo et al. \cite{39guo2022learning}. The AT can be formulated as the following minimax optimization problem \cite{GD-AT}
\begin{equation}
    \mathop{\arg\min}_{\theta} \mathop{\max}_{\tilde{\mathcal{G}} \in \mathcal{B}(\mathcal{G})} \sum^{n}_{i=1} \ell (f_\theta(\tilde{\mathcal{G}})_i, y_i)
\label{AT objective}
\end{equation}
where the inner maximization is to generate a perturbed graph $\tilde{\mathcal{G}}$ from the graph set $\mathcal{B(G)}$ with certain budgets of perturbations that the given clean graph $\mathcal{G}$ allows to generate, $f_\theta(\tilde{\mathcal{G}})_i$ is the prediction of node $i$ based on the perturbed $\tilde{\mathcal{G}}$ with the model parameters $\theta$, and $\ell$ is the loss of GNNs; the outer minimization consists of a normal GNN training process to find a robust model that produces correct label $y_i$ with the perturbed graph $\tilde{\mathcal{G}}$. Usually, the perturbed graph set is defined as $\mathcal{B(G)} \triangleq \{\tilde{\mathcal{G}}: \|A(\tilde{\mathcal{G}})\} - A(\mathcal{G})\|_0 \le \Delta\}$ where $A(\mathcal{G})$ is the adjacency matrix of $\mathcal{G}$, $\|A\|_0$ is the number of non-zero elements in the matrix $A$, and $\Delta$ specifies the budget of perturbed edges. 

With respect to AT for GNNs, in the transductive learning setting, both labeled and unlabeled data are available during training such that the model can employ self-training to achieve perfect robustness without sacrificing accuracy. This, however, makes the majority of the previous studies that focus on the setting of transductive learning misleading due to theoretical and empirical limitations pointed out by Lukas et al. in \cite{GD-AT}.
Therefore, this paper investigates \textbf{adversarial training in a fully inductive setting}, following the argument in \cite{GD-AT}. That is, in the inductive learning, the unlabeled data are completely unavailable during training, which is a more realistic scenario.

\section{Adversarial Training with Graph Subspace Energy (AT-GSE)}
\label{sec:AT-GSE}

In this section, we first introduce a new concept of graph subspace energy (GSE), followed by a novel adversarial training method with GSE being a regularization term. The GSE is used to produce worst-case perturbed graphs for adversarial training. To reduce time and space complexity, efficient GSE computation methods are also presented. 

\subsection{Graph Subspace Energy}
\label{sub:GSE}
\begin{definition}[Graph Subspace Energy]
    \label{def:GSE}
    For a graph $\mathcal{G}$ with the adjacency matrix $A \in \mathbb{R}^{n \times n}$, the graph subspace energy is defined as
    \begin{equation} \label{eq:GSE}
        \mathrm{GSE}(A,\beta_1,\beta_2) \triangleq \sum^{\lfloor\beta_2 n \rfloor}_{{i}=\lceil \beta_1 n \rceil} \sigma_i(A)
    \end{equation}
where $\beta_1<\beta_2 \in [0,1]$, $n$ is the number of nodes in the graph $\mathcal{G}$, $\sigma_i(A)$ is the $i$-th largest singular value of $A$ such that $\sigma_1(A) \geq \sigma_2(A) \geq \sigma_3(A) \ge...\geq \sigma_n(A) \ge 0$.
\end{definition}

Graph subspace energy is a generalization of the well-established concept of graph energy, which is a measure of graph structural properties related to the rank and the number of edges. With the singular values and the corresponding singular vectors, graph energy provides information about the connected shape and clustering structure of the graph.
From \Cref{def:GSE}, $\mathrm{GSE}(A,\beta_1,\beta_2)$ measures the sum of singular values from $k_1$ to $k_2$ with $k_1 = \lfloor\beta_1 n \rfloor < k_2=\lfloor\beta_2 n \rfloor$, which corresponds to a piece of singular value distribution of the adjacency matrix $A$. 
With the range $(k_1,k_2]$, $\mathrm{GSE}(A,\beta_1,\beta_2)$ could be able to take a specific care of the subspace spanned by the singular vectors associated with $\{\sigma_i\}_{i=k_1+1}^{k_2}$.  
When $\beta_1=0$ and $\beta_2=1$, $\mathrm{GSE}(A,0,1)$ reduces to nuclear norm.

\begin{figure}[!ht]
    \centering
    \subfloat[Cora]{\includegraphics[width=1.6in]{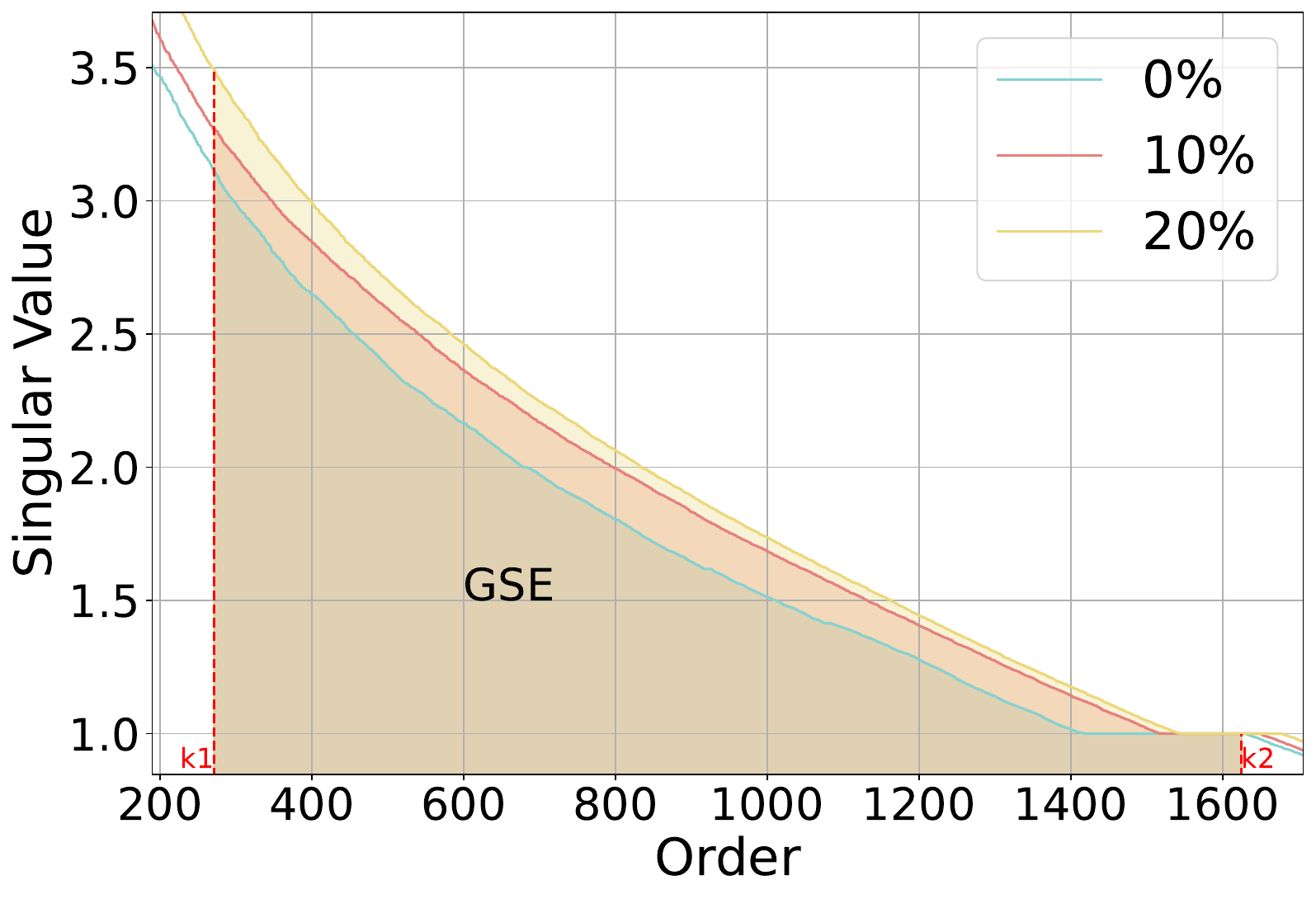}%
    \label{minifig:GSEarea_cora}}
    \hfil
    \subfloat[Cora]{\includegraphics[width=1.6in]{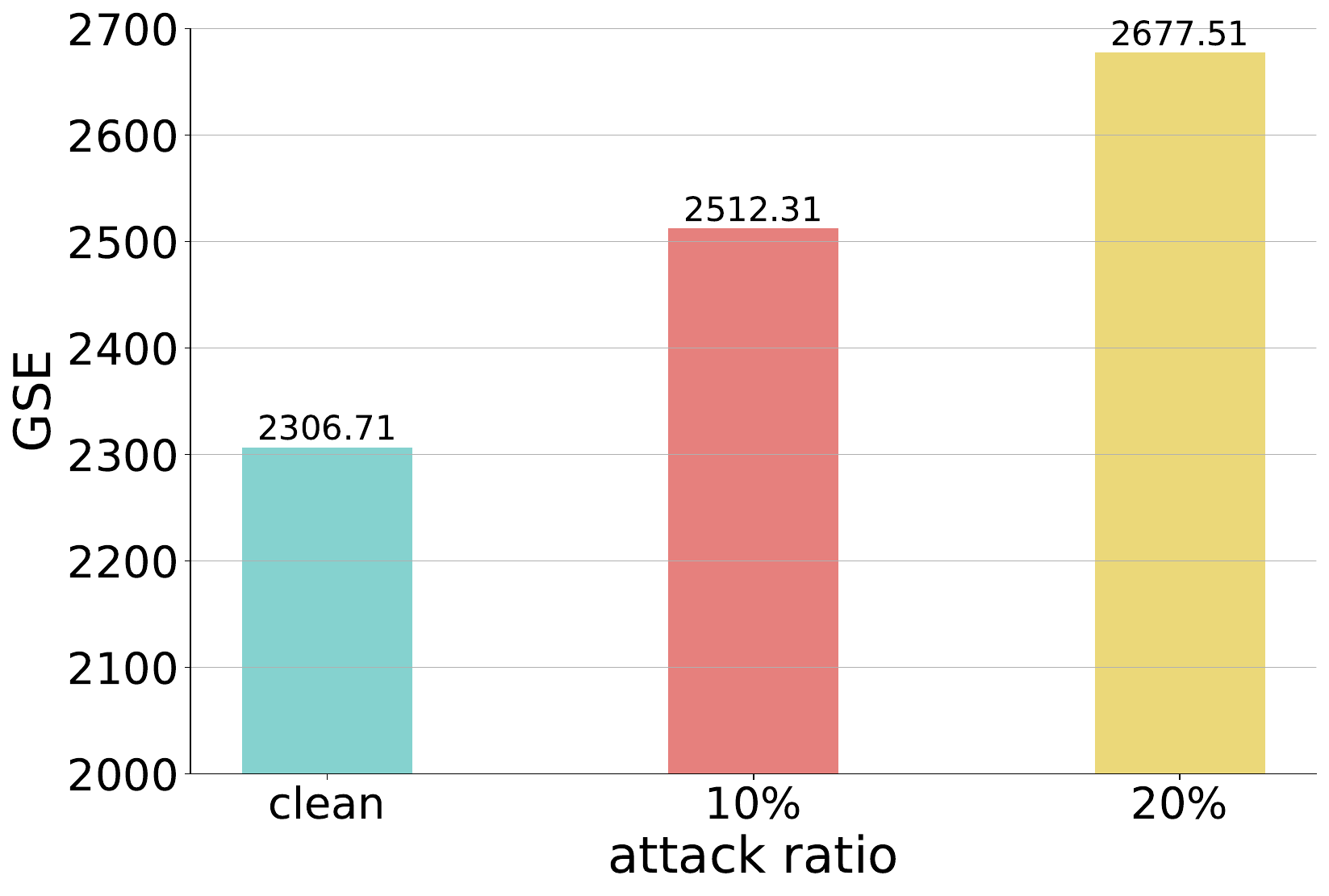}%
    \label{minifig:GSEbaro_cora}}
    \\
    \subfloat[Citeseer]{\includegraphics[width=1.6in]{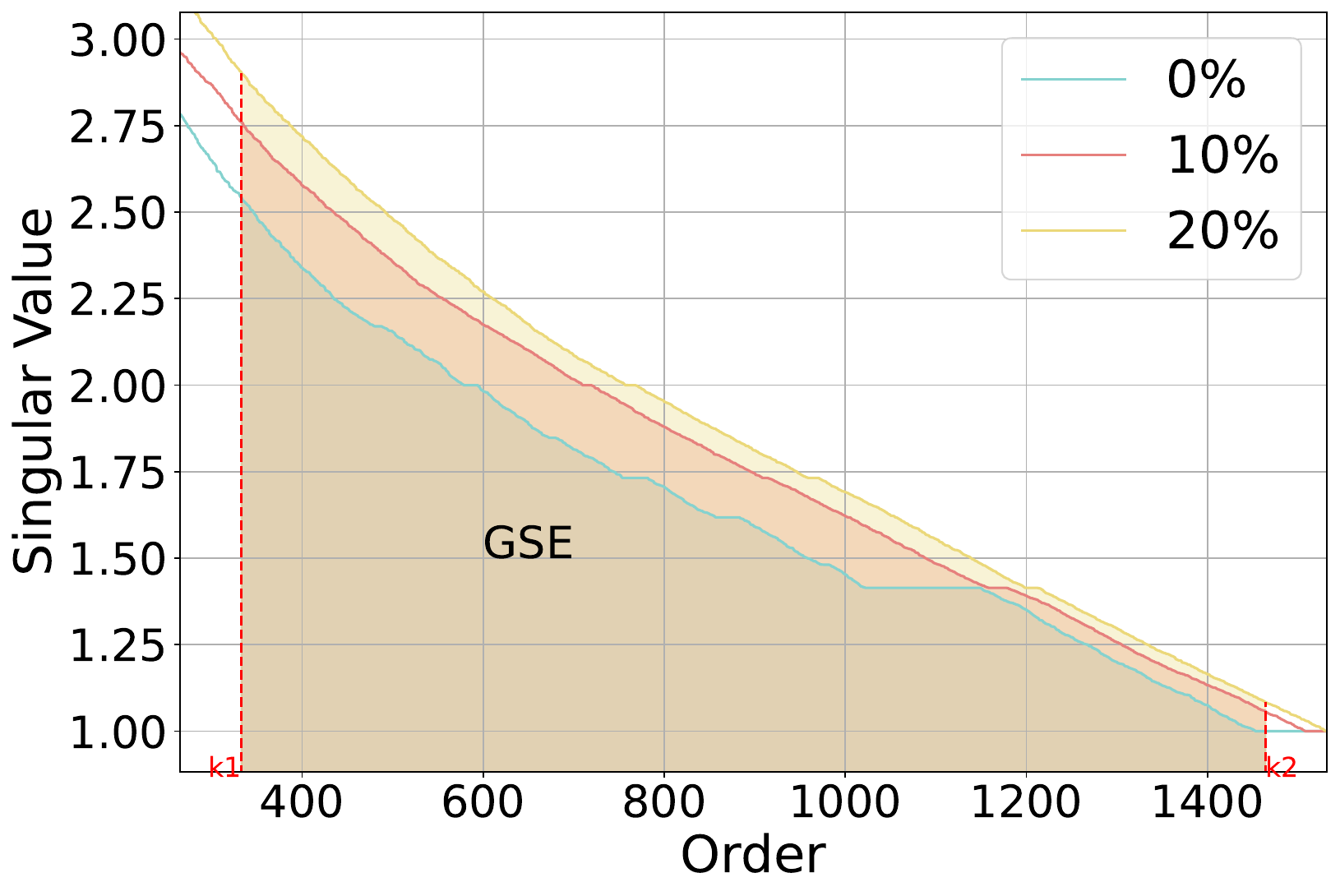}%
    \label{minifig:GSEarea_citeseer}}
    \hfil
    \subfloat[Citeseer]{\includegraphics[width=1.6in]{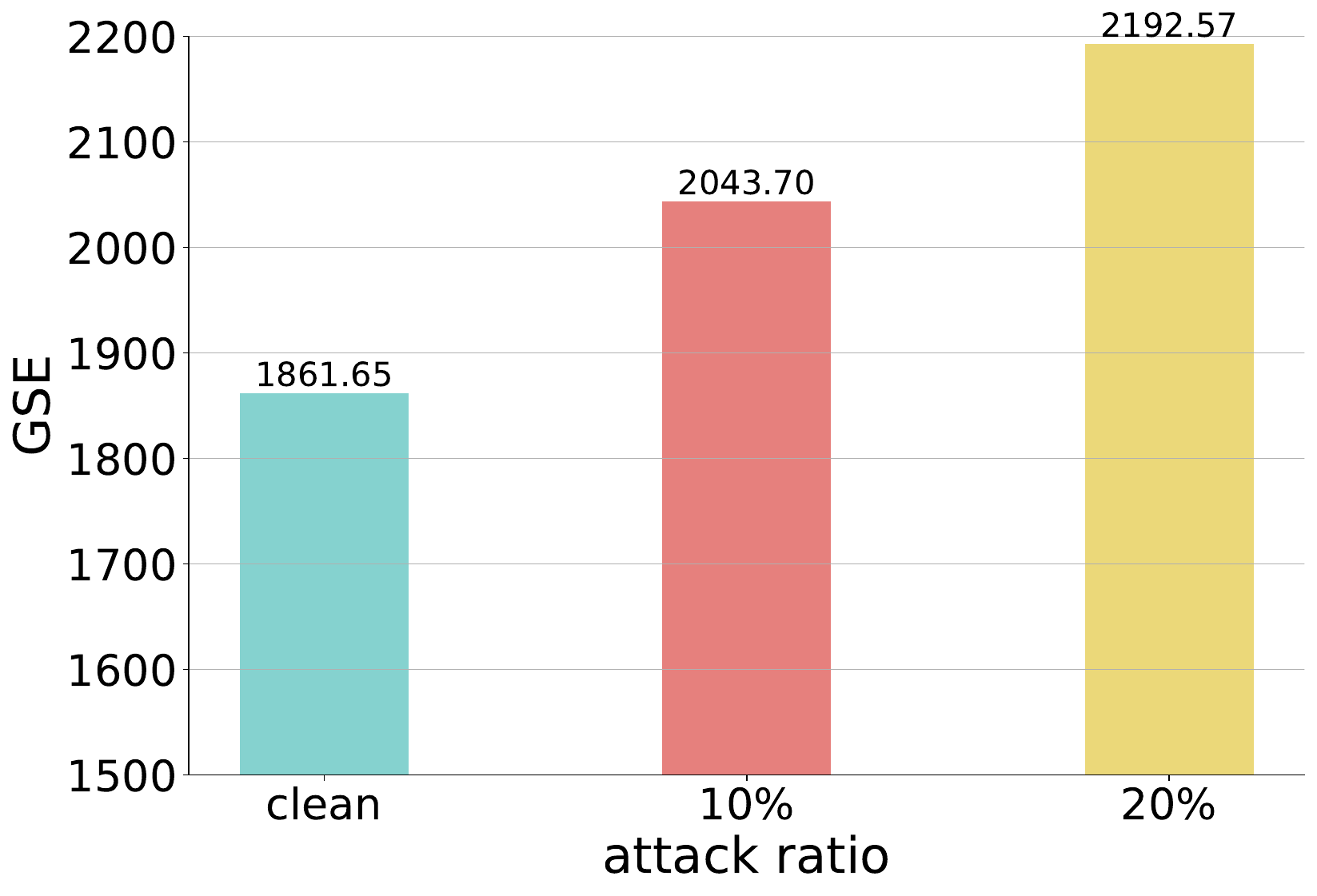}%
    \label{minifig:GSEbaro_citeseer}}
    \caption{A visual example of GSE under topology attacks of LRBCD with different attack ratios on Cora and Citeseer datasets.}
    \label{fig:GSE (lrbcd)}
\end{figure}

The GSE can be simply computed by performing singular value decomposition (SVD) on a matrix, where a set of singular values and their associated singular vectors are obtained. 
The GSE can be regarded as a generalized truncated SVD, which reserves only a range of singular value distribution and their corresponding vectors.  This truncation is commonly used for dimensionality reduction in various applications such as data compression, denoising, and low-rank approximation. It aims to preserve the essential information of the data in a lower dimensional subspace by retaining only the most important singular values.  So, the truncation method is used to try to obtain the graph subspace energy by truncating the graph energy.  We employ a more flexible truncation to freely retain the singular values according to the values of $k_1$ and $k_2$.  \Cref{fig:GSE (lrbcd)} visualizes the changed GSE under LRBCD attacks with different attack ratios on Cora and Citeseer. \Cref{minifig:GSEarea_cora} and \Cref{minifig:GSEarea_citeseer} on the left are the curves formed by the singular values of the original and perturbed adjacency matrices. The areas under the curves between $k_1$ and $k_2$ can be interpreted as the GSE specified by $\beta_1$ and $\beta_2$. The two right figures are GSE bar charts, each of which represents the  GSE value at the corresponding attack ratio. It can be clearly seen that as the attack ratio increases, the GSE increases.

\subsection{Singular Value Distribution Shift under Topology Attacks}
\label{sub:SVD}
Adversarial training (AT) in GNNs is to improve the robustness and generalization capabilities of models against adversarial attacks or perturbations by exposing the learning model to adversarial examples during training. Following \cite{GD-AT}, we consider AT in a fully inductive learning.
Adversarial examples are small perturbations injected into the input data on purpose in order to cause the model to misclassify the resulting perturbed data points. Searching for potential adversarial examples is usually an efficient way to identify the vulnerability of the trained models. By looking into the specific adversarial examples, it is hoped to find the possible underlying patterns. 
\begin{figure}[!ht]
    \centering
    \subfloat[Cora]{\includegraphics[width=1.6in]{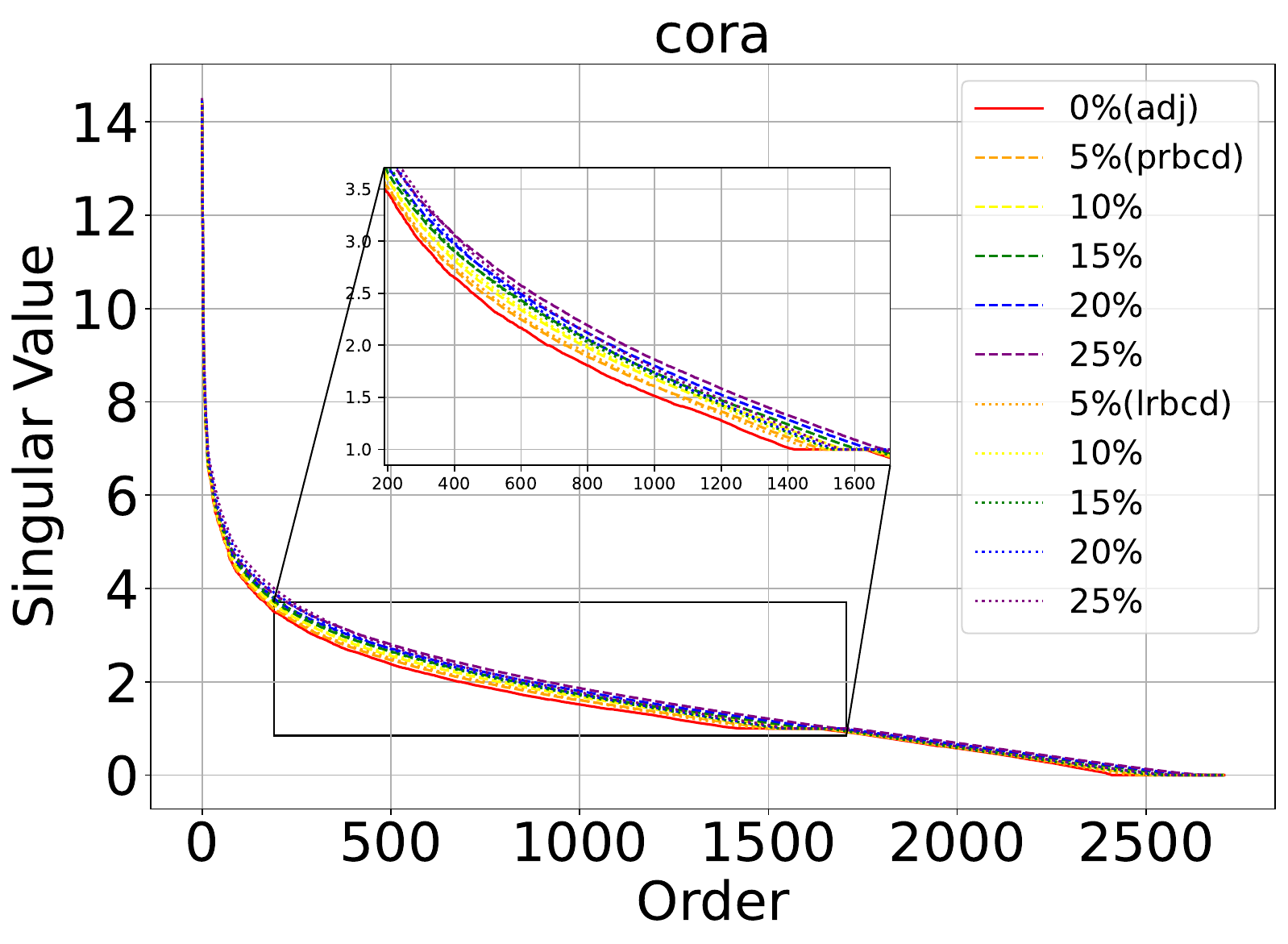}%
    \label{minifig:singular_cora}}
    \hfil
    \subfloat[Citeseer]{\includegraphics[width=1.6in]{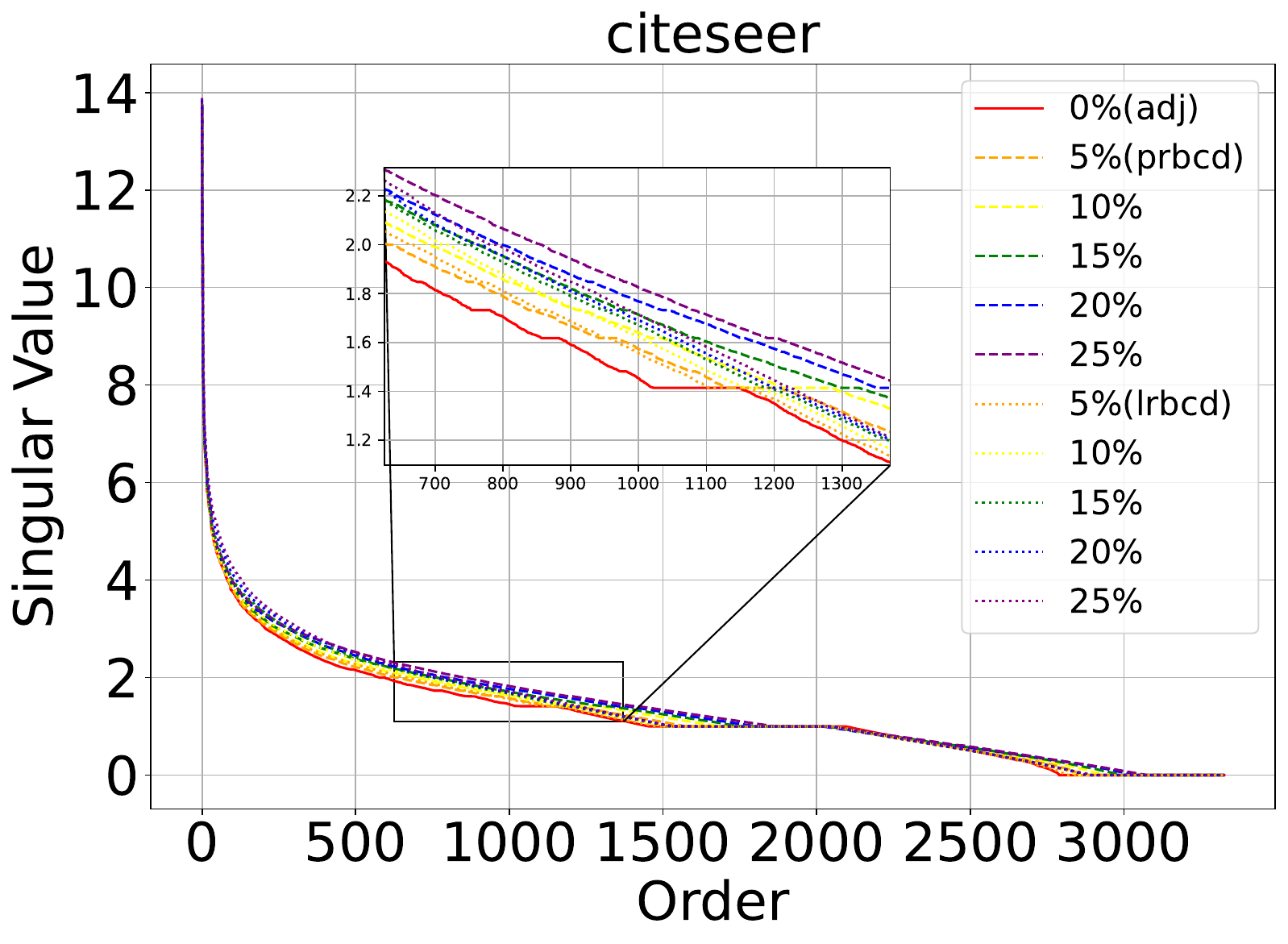}%
    \label{minifig:singular_citeseer}}
    \caption{The distribution shift of singular values of graph adjacency matrix under topology attacks of PRBCD and LRBCD with different attack ratios on Cora and Citeseer datasets.}
    \label{fig:singular values (bcd)}
\end{figure}

\Cref{fig:singular values (bcd)} showcases the singular value distribution shift of the adjacency matrices of the Cora and Citeseer datasets under different topology attacks (i.e., PRBCD and LRBCD), where there is a right-shift within a range of singular values. Therefore, the graph subspace energy (GSE) has the potential to have an influence on measuring the attack levels, i.e., the ratios of the number of perturbed edges to the total number of edges. Specifically, adversarial examples intend to change singular values of the adjacency matrices in a specific subspace.

\subsection{GSE Offset Attacks and Adversarial Training}
\label{sub:GSE_offset}

The above observation implies that adversarial topology attacks result in a potential right-shift of singular value distribution of the adjacency matrix, corresponding to an increase of the GSE of the associated graph. Inspired by this observation, in addition to the GNN adversarial training in \Cref{AT objective}, the graph subspace energy term is added to the adversarial training as a regularizer to produce the adversarially perturbed graph. This framework is referred to as adversarial training with graph subspace energy (AT-GSE), whose minimax optimization problem can be formulated as

\begin{equation}
    \mathop{\arg\min}_{\theta} \max_{\tilde{A} \in \mathcal{B}(A)} \left\{ \mathcal{L}_{GNN}(\theta, \tilde{A}, X, Y) + \gamma 
  \mathrm{GSE} (\tilde{A},\beta_1,\beta_2) \right\}
\label{eq:AT-GSE_obj}
\end{equation}
where $ \mathcal{B}(A) \triangleq \{\tilde{A}: \|A - \tilde{A}\|_0 \le \Delta\}$ is the set of adjacency matrices of the perturbed graphs $\tilde{\mathcal{G}}$ with the number of perturbed edge budget being bounded by $\Delta$, $\mathcal{L}_{GNN}(\theta, A, X, Y)$ is the expected/empirical loss of the GNN model with model parameter $\theta$, adjacency matrix $A$, node attribute/feature set $X$ and label set $Y$,
$\mathrm{GSE}$ is the regularization term to help produce perturbed graph $\tilde{A}$ with maximized GSE, and $\gamma$ is a predefined hyper-parameter to balance between the adversarial loss and the GSE regularization term.

The AT-GSE minimax optimization problem consists of an inner maximization problem to find a worst-case topology perturbation that increases the GNN loss function and the GSE term within the perturbed graph set, and an outer minimization is to obtain a robust GNN model to minimize the GNN loss given the perturbed graph. While the outer minimization can be solved alternatively by stochastic gradient descent, the inner maximization to find the worst-case perturbed graph is more challenging due to the gradient computation of the GSE term.

Regarding the inner maximization as to finding a GSE topology attack, we have two ways to generate the worst-case perturbed graph with adjacency matrix $\tilde{A}$. One simpler way is to randomly generate a number of perturbed graphs $\tilde{A}$ within the graph set of $\mathcal{B}(A)$ in such a way that the perturbed graph with the maximum GSE is selected. This way of graph perturbation is referred to as ``RndGSE'' attack. On the other hand, given the above observation of singular value distribution shift, we introduce a GSE offset attack that aims to increase the singular values of the adjacency matrix $A$ by simply adding a constant $\alpha$ to the singular values of interest, i.e.,
\begin{equation}
\label{eq:GSE_offset}
\mathrm{GSE}_{\alpha}(A,\beta_1,\beta_2) = \sum^{{k_2}}_{{i=k_1+1}} \big(\sigma_i(A) + \alpha \big)
\end{equation}
corresponding to a right-shift of singular value distribution, where $\alpha$ is a hyper-parameter fine-tuned with $\beta_1$ and $\beta_2$.
Given the perturbation budget $\|A-\tilde{A}\|_0 \le \Delta$, we have a reference value of $\alpha$.
Let $\tilde{A}=A+E$ where $E \in \{0,1\}^{n \times n}$ is the perturbation matrix such that $\|E\|_0 \le \Delta$.
According to the Hoffman-Wielandt inequality \cite{drineas2005nystrom}
\begin{equation}
\label{eq:Nystrom_sumalpha}
    \sum^n_{k=1} \left( \sigma_k(A+E) - \sigma_k(A) \right)^2 \le \| E \|^2_F \le \Delta,
\end{equation}
we can roughly set $ n \alpha^2 \le \Delta$ to ensure a reasonable adjustment, i.e., $\alpha \leq \sqrt{\frac{\Delta}{n}}$.

Therefore, the inner maximization can be done by gradient ascent of the GNN loss of the perturbed graph $\tilde{A}$, together with the additional gradient of the GSE regularizer.
To obtain the gradient of GSE, we extend the proximal operator in \cite{proximal-operator} to the derivative of GSE offset. The projection follows the generation rules of $\mathrm{GSE}_{\alpha}$, increases the value of the truncated part of the singular values, and retains the largest singular value while discarding the smallest singular values, so as to produce the perturbed adjacency matrix. 
Specifically, the proximal operator on the GSE term can be written by
\begin{multline}
    \label{eq:GSE_operator}
    \mathrm{prox}_{\alpha \|.\|_{(k)}}(Z_k) = U_{k_2} \mathrm{diag}\big(\sigma_1, \dots, \sigma_{k_1}, \\
    \quad (\sigma_{k_1+1} + \alpha), \dots, (\sigma_{k_2} + \alpha) \big) V_{k_2}^T
\end{multline}
where  $k$ is interpreted as $k_1=\lfloor\beta_1 n \rfloor$ and $k_2=\lfloor\beta_2 n \rfloor$, $0 \leq \beta_1 < \beta_2 \leq 1$, which are predefined parameters to represent the proportion, $Z = U \mathrm{diag}(\sigma_1,...,\sigma_n) V^T$ is the singular value decomposition of $Z$, and $U_{k_2}, V_{k_2}$ are the concatenated singular vectors associated with the singular values $\{\sigma_i\}_{i=1}^{k_2}$. It is worth noting that only the first $k_2$-dimensional subspace is selected to generate the perturbed graph and the last $k_2-k_1$ singular values are adjusted with a constant $\alpha$. 

\begin{algorithm}[!ht]
    \caption{AT-GSE Algorithm}
    \label{alg:AT-GSE}
    \DontPrintSemicolon
    \BlankLine
    \KwIn{Training/validation graph $\mathcal{G}_{t/v}$, training/validation adjacency $A_{t/v}$, training/validation labels $y_{t/v}$, GNN $f_{\theta_0}$, epochs $E$, warm-up epochs $W$, loss $\ell$, learning rate $\eta$,  Hyper-parameters $\alpha, \beta_1, \beta_2$}
    \KwOut{GNN $f_{\theta^*}$}
    Initialize $\ell_{min} \gets \infty$ \; 
    \For{$l = 1$ \KwTo $W$}
    {
        $\theta_l \gets \theta_{l-1} + \eta \nabla_{\theta_{l-1}}(\ell(f_{\theta_{l-1}}(\hat{\mathcal{G}}_t),y_t)$
    }
    \For{$l=W$ \KwTo $E$}
    {
        \While{stopping condition is not met}
        {
            $\tilde{A}_t \gets \tilde{A}_t + \eta \nabla_{\tilde{A}_t}(\mathcal{L}_{GNN})$ \;
            $\tilde{A}_t \gets \mathrm{prox}_{\alpha \|.\|_{(k)}}(\tilde{A}_t) \longleftarrow$ \Cref{eq:GSE_operator}  \;    
        }
        $\hat{\mathcal{G}}_t \gets \tilde{A}_t$ \;
        $\theta_l \gets \theta_{l-1} + \eta \nabla_{\theta_{l-1}}(\ell(f_{\theta_{l-1}}(\hat{\mathcal{G}}_t),y_t)$ \;
        \While{stopping condition is not met}
        {
            $\tilde{A}_v \gets \tilde{A}_v + \eta \nabla_{\tilde{A}_v}(\mathcal{L}_{GNN})$ \;
            $\tilde{A}_v \gets \mathrm{prox}_{\alpha \|.\|_{(k)}}(\tilde{A}_v) \longleftarrow$ \Cref{eq:GSE_operator}  \;    
        }
        $\hat{\mathcal{G}}_v \gets \tilde{A}_v$ \;
        \If{$\ell_{min} > \ell(f_{\theta_l}(\hat{\mathcal{G}}_v),y_v)$}
        {
            $\ell_{min} \gets \ell(f_{\theta_l}(\hat{\mathcal{G}}_v),y_v)$ \;
            $\theta^* \gets \theta_l$
        }
    }
\end{algorithm}

Using these updates and the projection rules, \Cref{alg:AT-GSE} summarizes the details of our proposed AT-GSE method. Note that gradient ascent is employed to maximize the GNN loss function, and GSE offset, i.e., $\mathrm{GSE}_{\alpha}$, is used to enhance the graph subspace energy, both of which aim to produce perturbed graphs for outer minimization.

According to the proximal operator of \Cref{eq:GSE_operator}, we can clearly find that for GSE offset, how to effectively find the singular values and singular vectors of the adjacency matrix is the key point of the algorithm. The most common and original method is the SVD operation by, i.e., directly calculating the singular value matrix and the left and right singular value vector matrix of the adjacency matrix, and then performing related operations on them to get the final modified matrix.

\subsection{Efficient GSE Computations}
While directly employing SVD can get accurate singular values for our algorithm, the computational complexity of SVD is high, which may lead to a huge computing time for large-scale graph datasets. Moreover, for sparse graphs, directly applying SVD to adjacency matrices will produce three dense matrices, which may lead to a large increase in storage overhead. Therefore, although SVD has wide applicability in theory, it may encounter some challenges in practical applications, especially when dealing with large-scale and sparse graphs. In order to handle large-scale and sparse graph datasets, we resort to the randomized singular value decomposition (RndSVD), and Nystr\"om low-rank approximation methods.

\subsubsection{Randomized SVD (RndSVD)}
\label{sub:RndSVD}
The core idea of RndSVD based AT-GSE is to generate adversarial examples that approximates the SVD of large matrices through random projection and dimensionality reduction techniques. This method is particularly suitable for processing very large and sparse matrices. It is roughly divided into three main steps: constructing a low-dimensional submatrix (line 8-14), calculating the SVD of the low-dimensional matrix (line 15-16), and constructing a perturbed adjacency matrix by the decomposed values. The detailed method is shown in \Cref{alg:AT-RndSVD}.

\begin{algorithm}[!ht]
    \caption{AT-GSE by RndSVD with Power Iteration}
    \label{alg:AT-RndSVD}
    \DontPrintSemicolon
    \BlankLine
    \KwIn{Graph $\mathcal{G}$, adjacency $A$, labels $y$, GNN $f_{\theta_0}$, epochs $E$, warm-up epochs $W$, loss $\ell$, learning rate $\eta$,  Hyper-parameters $\alpha$, $0 \leq \beta_1 < \beta_2 \leq 1$, number of power iterations $q$}
    \KwOut{GNN $f_{\theta^*}$}
    Initialize $\ell_{min} \gets \infty$, rank parameter $k_1=\lfloor\beta_1 n \rfloor$, $k_2=\lfloor\beta_2 n \rfloor$ \;
    \For{$l = 1$ \KwTo $W$}
    {
        $\theta_l \gets \theta_{l-1} + \eta \nabla_{\theta_{l-1}}(\ell(f_{\theta_{l-1}}(\hat{\mathcal{G}}),y)$
    }
    \For{$l=W$ \KwTo $E$}
    {
        \While{stopping condition is not met}
        {
            $\tilde{A}_t \gets \tilde{A} + \eta \nabla_{\tilde{A}}(\mathcal{L}_{GNN})$ \;
            $\Omega \in \mathbb{R}^{n \times (k_2+p)} \gets$ randn$\big (n,(k_2+p) \big)$ \;
            $Y = \tilde{A} \Omega \in \mathbb{R}^{n \times (k_2+p)} $ \;
            \For{$j = 1$ \KwTo $q$}
            {
                $Y \gets \tilde{A} (\tilde{A}^T Y)$ \;
            }
            $Q \in \mathbb{R}^{n \times (k_2+p)} \gets$ qr$(Y)$ \;
            $B \in \mathbb{R}^{(k_2+p) \times n} \gets Q^T \tilde{A}$ \;
            $U_B, \Sigma, V \gets$ svd$(B)$ \;
            $U \gets QU_B$ \;
            $\tilde{A} \gets \mathrm{prox}_{\alpha \|.\|_{(k)}}(U,\Sigma,V) \longleftarrow$ \Cref{eq:GSE_operator}  \;
        }
        $\hat{\mathcal{G}} \gets \tilde{A}$ \;
        \If{$\ell_{min} > \ell(f_{\theta_l}(\hat{\mathcal{G}}),y)$}
        {
            $\ell_{min} \gets \ell(f_{\theta_l}(\hat{\mathcal{G}}),y)$ \;
            $\theta^* \gets \theta_l$
        }
    }
\end{algorithm}

The algorithm approximates the SVD of the matrix through random projection instead of exact calculation. 
Give a matrix $A$ and a random tall matrix $\Omega$, RndSVD computes the SVD on a low-dimensional matrix, which is the random projection of $A$ onto the subspace spanned by the sampling matrix $\Omega$.
According to \cite{halko2011finding_RndSVD}, the approximation error satisfies 
\begin{equation}
    \label{eq:ARndSVD_error}
    \|A - \tilde{Q}\tilde{Q}^{T}A\| \leq \left[ 1 + 9 \sqrt{k+p} \cdot \sqrt{n} \right] \sigma_{k+1}
\end{equation}
with probability at least $1-3 \cdot p^{-p}$,   
where $\tilde{Q}\tilde{Q}^{T}$ is the subspace spanned by the random projection, $k$ is the target rank, $p$ is the oversampling parameter, $k+p \leq n$, and $\sigma_{k+1}$ denotes the $(k+1)$-th largest singular values of $A$. 
When the oversampling $p = 5$, the failure probability of \Cref{eq:ARndSVD_error} is $3 \cdot 5 ^{-5} = 0.00096$. That is a very small probability, indicating that the reliability of the approximate result is very high, thereby 5 is a reasonable assumption for oversampling.

In order to obtain a more accurate approximation result in randomized SVD, we combine the power iteration method to update $Y$, where the number of iterations is $q$, as shown in line 10-12 of \Cref{alg:AT-RndSVD}. Hence the expected approximate error bound \cite{halko2011finding_RndSVD} with $q$ is written as 
\begin{equation}
    \label{eq:RndSVD_power}
    \mathbb{E}\|A - U \Sigma V^T\| \leq \left [ 1 + 4\sqrt{\frac{2n}{k-1}}\right]^{1/2q+1} \sigma_{k+1}.
\end{equation}
Notably, as $q$ increases, the error becomes smaller. That is, the more iterations there are, the smaller the approximation error is. Certainly, the more iterations, the longer the running time. So we also need to determine a reasonable value of $q$ through the following time complexity.

RndSVD reduces computational complexity by compressing large matrices into smaller ones through random projection. The time complexity of RndSVD with power iteration is \cite{halko2011finding_RndSVD}
\begin{equation}
    \label{eq:RndSVD_time}
    T_{RndSVD} = (2q+2)kT_{mult} + O(2nk^2)
\end{equation}
where $T_{mult}$ is the ﬂop count of a matrix-vector multiply with $A$.
RndSVD is particularly suitable for large matrices that cannot be fully stored in memory. 
Only small matrices are calculated, which reduces memory usage. This is particularly important for processing large datasets, especially when the dataset is too large to be fully loaded into memory and the traditional SVD algorithm cannot be used. 
\Cref{eq:RndSVD_time} suggests that the time complexity is closely related to $q$. In practice, in order to ensure low time and space complexity, $q=1$ or $q=2$ is sufficient.

Overall, RndSVD provides an effective solution for large-scale datasets, which may sacrifice some accuracy in exchange for computational efficiency and memory usage efficiency. Although the traditional SVD algorithm can theoretically provide an accurate decomposition, it may encounter memory and computing time limitations on large graphs.

\subsubsection{Nystr\"om Method}
\label{sub:Nystrom}

\begin{algorithm}[!ht]
    \caption{AT-GSE by Nystr\"om Method}
    \label{alg:Nystrom}
    \DontPrintSemicolon
    \BlankLine
    \KwIn{Graph $\mathcal{G}$, adjacency $A$, labels $y$, GNN $f_{\theta_0}$, epochs $E$, warm-up epochs $W$, loss $\ell$, learning rate $\eta$,  Hyper-parameters $\alpha$, $0 \leq \beta \leq 1$}
    \KwOut{GNN $f_{\theta^*}$}
    Initialize $\ell_{min} \gets \infty$, rank parameter $k_2=k=\lfloor\beta n \rfloor$ \;
    \For{$l = 1$ \KwTo $W$}
    {
        $\theta_l \gets \theta_{l-1} + \eta \nabla_{\theta_{l-1}}(\ell(f_{\theta_{l-1}}(\hat{\mathcal{G}}),y)$
    }
    \For{$l=W$ \KwTo $E$}
    {
        \While{stopping condition is not met}
        {
            $\tilde{A}_t \gets \tilde{A} + \eta \nabla_{\tilde{A}}(\mathcal{L}_{GNN})$ \;
             $c \gets \text{randperm}(n)[:k]$ \;
            $C \in \mathbb{R}^{n \times k} \gets \tilde{A}[:, c]$ \;
            $W \in \mathbb{R}^{k \times k} \gets C[c, :]$ \;
            $W^{\dagger} \gets \text{pinv}(W)$ \;
            $\hat{A} \gets C  W^{\dagger}  C^T$ \;
            $\tilde{A} \gets \alpha \hat{A}$ \;
        }
        $\hat{\mathcal{G}} \gets \tilde{A}$ \;
        \If{$\ell_{min} > \ell(f_{\theta_l}(\hat{\mathcal{G}}),y)$}
        {
            $\ell_{min} \gets \ell(f_{\theta_l}(\hat{\mathcal{G}}),y)$ \;
            $\theta^* \gets \theta_l$
        }
    }
\end{algorithm}

The Nystr\"om method  reduces high time and space complexity by generating low-rank matrix approximations. 
It works as an information fusion method by using parts of the data multiple times to approximate the values we are interested in, such as the eigenvalues/eigenvectors of a matrix or the inverse of a matrix. Applying the Nystr\"om method to machine learning problems can improve efficiency without significantly reducing performance. 

Therefore, in order to further reduce the time complexity, we enhance AT-GSE with Nystr\"om, by using it to find a low-rank approximation matrix, and then multiply it by a hyper-parameter $\alpha$ to generate a perturbed graph, thereby constructing an adversarial sample. The AT-GSE with Nystr\"om method appears in \Cref{alg:Nystrom}. Specifically, the Nystr\"om method only needs to store and process a small part of the data $W$. It generates a low-rank approximate matrix $\hat{A}$ by using $W$ many times, which can improve efficiency without significantly reducing performance.

The approximation error of the Nystr\"om method under appropriate assumptions can be written \cite{drineas2005nystrom}
\begin{equation}
 \label{eq:ANystrom_error}
    \|A - C  W^{\dagger}  C^T\|_\xi \leq \|A - A_k\|_\xi + \epsilon \sum^n_{k=1} A_{ii}^2
\end{equation}
where $\xi = 2, F$, $\|\cdot\|_2$ and $\|\cdot\|_F$ denote the $\ell_2$ and the Frobenius norm, respectively. And $A_k$ is the best $k$-rank approximation of $A$, $\epsilon$ is an error parameter.
\Cref{eq:ANystrom_error} shows that the error between the approximation matrix $\hat{A}$ and the original matrix $A$ will not exceed the error of the best $k$-rank approximation $A_k$ plus an additional term with an adjustable parameter $\epsilon$.

In conclusion, AT-GSE with the Nystr\"om low-rank approximation focuses on the low-frequency band of graph spectrum, paying more attention to global topology changes. It is evident from Nystr\"om's implementation steps that the largest $k_2$ singular values are adjusted. It is different from AT-GSE with RndSVD that preserves the low-frequency part corresponding to the largest $k_1 < k_2$ singular values whilst only adjusts the graph spectrum from $k_1$ to $k_2$.
Therefore, compared with AT-GSE with RndSVD emphasizes more on local topology changes, the Nystr\"om method may pay more attention to global changes. Such differences make AT-GSE with RndSVD and Nystr\"om method suitable respectively to defending against the local attacks (i.e., LRBCD) and the global attacks (i.e., PRBCD), as showcased in the experiments.

\section{Experiments}
\label{sec:experiments}
\textit{Datasets.}
Cora, Cora\_ML, Citeseer \cite{homophily(cora-citeseer)}, and Pubmed \cite{pubmed} datasets are commonly used in the fields of graph semi-supervised learning for citation network analysis. All three datasets are citation networks, where nodes represent articles and edges represent citation relationships. These datasets are commonly used for tasks such as node classification and link prediction. 
The SBM (stochastic block model) \cite{SBM} datasets are used in the field of network analysis and community detection. This dataset contains synthetic rather than real-world data, and the node features are Gaussian distributed. The adjacency matrices of SBM graphs exhibit a block structure, where nodes within the same block are more densely connected than those in different blocks.
WikiCS is a graph dataset based on Wikipedia. The nodes correspond to Wikipedia papers in the field of computer science, and the edges are constructed based on the hyperlinks between papers. It also contains 10 categories representing different branches in the field of computer science.
The statistics of these datasets are shown in \Cref{table:datasets}.

\begin{table}[!ht]\centering
\renewcommand\arraystretch{1.2}
\caption{Dataset statistics}
\scalebox{0.85}{
\begin{tabular}{l|ccccc|cc}
    \hline
    & \multicolumn{5}{c|}{Homophily} & \multicolumn{2}{c}{Heterophily} \\ \hline
    Datasets & Cora & Citeseer & Cora\_ML & Pubmed & SBM & SBM & WikiCS \\ 
    \hline
    \#Nodes & 2708 & 2110 & 2810 & 19717 & 981 & 980 & 10311 \\ 
    \#Edges & 10556 & 7336 & 15962 & 44338 & 3890 & 3956 & 431108 \\
    \#Features & 1433 & 3703 & 2879 & 500 & 21 & 21 & 300 \\ 
    \#Classes & 7 & 6 & 7 & 3 & 2 & 2 & 10 \\ 
    \hline
\end{tabular}}
\label{table:datasets}
\end{table}

\textit{Inductive Learning.} 
Our focus is placed on inductive graph semi-supervised learning, where the validation and test nodes are not available during training so that merely memorizing the clean graphs does not lead to perfect robustness \cite{GD-AT}. 
All experiments in \Cref{sec:experiments} follow a fully inductive semi-supervised setting as in \cite{GD-AT}. 

\textit{Baselines.}
To evaluate the effectiveness of our proposed AT-GSE method on node classification, we compare it with some classical GNN models with and without adversarial training (AT). These baselines are briefly introduced as follows.
\begin{itemize}
    \item[$\bullet$] \textbf{GCN \cite{GCN}} is the cornerstone of most graph neural networks, but its lack of robustness makes it extremely vulnerable to adversarial attacks.
    \item[$\bullet$] \textbf{APPNP \cite{APPNP}} combines personalized propagation with neural prediction, where information from neighborhoods is propagated in a personalized way, giving more weight to nodes that are more relevant to the target nodes to achieve accurate and scalable predictions.
    \item[$\bullet$] \textbf{GPRGNN \cite{GPRGNN}:} generalized pagerank (GPR) can adaptively learn the weight of GPR, and then automatically adjust to the node label mode, thereby jointly optimizing the extraction of node features and topological information, ensuring that the model maintains excellent learning performance regardless of whether the node label is homophilic or heterophilic.
    \item[$\bullet$] \textbf{GNNs with AT \cite{GD-AT}:} The above standard GNNs can be trained with adversarial training, by employing adversarial perturbations (e.g., LRBCD and PRBCD attacks) to generate adversarial examples for training data augmentation.
\end{itemize}

\textit{Setup.}
In the induction experiments of this paper, 20 nodes of each class were sampled as labeled training and validation sets. In addition, the test set consists of 10\% of all nodes, and the remaining nodes are used as additional unlabeled nodes to be added to the training and validation sets. As mentioned in the baseline above, the comparison models in the following experiments include GCN, APPNP, GPRGNN, and these models with adversarial training. All experimental results are the mean accuracy and standard deviation obtained after running the model 10 times. Our method employs GCN, APPNP, and GPRGNN as the backbone GNN models to find the maximal loss function in the experiments. By adjusting the truncated singular values of the adjacency matrix, through modifying $\alpha$ and $\beta$ in \Cref{eq:AT-GSE_obj}, we obtain the maximized GSE term, thereby generating the expected robust model.

\subsection{Hypothesis}
Our experiments aim to verify the following hypotheses: 
\begin{itemize}
    \item[$1).$] AT-GSE can improve over the natural training on existing GNN models with respect to the clean accuracy.
    \item[$2).$] AT-GSE can improve adversarial robustness on different datasets, including homophilic and heterophilic graphs, improving over state-of-the-art methods.
    \item[$3).$] Low-rank approximation methods (Nystr\"om) are more suitable for defending against global attacks -- PRBCD, while more specific spectrum adjustments (RndSVD) are more suitable against local attacks -- LRBCD.
\end{itemize}

\subsection{Robustness Performance}
Our focus on robustness performance is placed on the ability of the GNN models to maintain accuracy against adversarial topology attacks, or perturbations, referred to as adversarial accuracy. 
To empirically verify the effectiveness of our proposed AT-GSE method, we evaluate the adversarial accuracy of the trained GNN models with different AT methods against topology perturbations on homophily and heterophily datasets under different ratios of perturbed edges. 
The following tables report the node classification results of AT-GSE and its related models, and some classic GNNs with and without adversarial training are included for comparison. From the results, we have the following observations.

\textbf{1) Models with adversarial training generally achieve better adversarial robustness than their original models.}
\Cref{table:AT_attacks} shows that adversarial training based on multiple backbone models has a certain defensive effect against various attacks. Specifically, adversarial training helps improve the defense capabilities of GNNs against malicious attacks and adversarial threats. By training the network to withstand adversarial examples, the impact of small budgets on the input data can be mitigated, thereby improving the reliability of the network. 

\begin{table}[!ht]
\renewcommand\arraystretch{1.2} 
\caption{Comparison between AT-GSE and baselines under various attacks on Citeseer without self-training. 
The first column (Attack), second column (Model) and third column (AT, × means models without AT) show the names of the attacks, baseline models, and our proposed methods (AT-GSE) together. APPNP and GPRGNN represent the backbone GNN model. The percentage (0\%, 10\%, and 25\%) is the rate of budget. The best result is highlighted in \textbf{bold}.}
\scalebox{0.91}{
\centering
\begin{tabular}{c|c|c|c|c|c}
    \hline
    Attack & Model & AT & 0\% & 10\% & 25\% \\
    \hline
    \multirow{6}{*}{LRBCD} & \multirow{3}{*}{APPNP} & × & 70.93±0.88 & 62.24±1.18 & 56.03±1.58 \\
    & & LRBCD & 71.96±0.00 & 63.04±0.82 & 57.76±1.70 \\
    & & AT-GSE & \textbf{73.97±0.30} & \textbf{68.46±1.28} & \textbf{65.23±0.94} \\
    \cline{2-6}
    & \multirow{3}{*}{GPRGNN} & × & 59.63±0.84 & 54.07±0.91 & 45.05±1.19 \\
    & & LRBCD & 59.35±0.00 & 56.45±0.65 & 53.79±1.15 \\
    & & AT-GSE & \textbf{72.90±0.00} & \textbf{71.45±0.25} & \textbf{67.85±0.72} \\
    \hline
    \multirow{6}{*}{PRBCD} & \multirow{3}{*}{APPNP} & × & 70.75±1.29 & 59.58±1.33 & 50.84±1.49 \\
    & & LRBCD & \textbf{73.55±0.56} & 62.52±0.50 & 54.35±0.91 \\
    & & AT-GSE & 73.22±0.51 & \textbf{65.75±0.81} & \textbf{60.05±0.64} \\
    \cline{2-6}
    & \multirow{3}{*}{GPRGNN} & × & 60.28±2.80 & 53.46±1.76 & 47.38±1.39 \\
    & & LRBCD & 58.97±5.67 & 55.79±2.79 & 53.18±1.60 \\
    & & AT-GSE & \textbf{71.64±1.18} & \textbf{69.67±0.64} & \textbf{68.50±1.05} \\
    \hline
    \multirow{6}{*}{PGD} & \multirow{3}{*}{APPNP} & × & 69.72±1.12 & 63.46±1.28 & 55.65±0.53 \\
    & & LRBCD & 71.68±0.84 & 63.22±0.98 & 57.52±1.26 \\
    & & AT-GSE & \textbf{73.46±0.65} & \textbf{67.76±1.02} & \textbf{62.85±0.82} \\
    \cline{2-6}
    & \multirow{3}{*}{GPRGNN} & × & 63.55±1.14 & 56.92±0.69 & 52.24±1.83 \\
    & & LRBCD & 62.34±1.50 & 60.00±1.50 & 56.92±0.19 \\
    & & AT-GSE & \textbf{74.25±0.92} & \textbf{71.96±0.72} & \textbf{68.83±0.21} \\ \hline
\end{tabular}}
\label{table:AT_attacks}
\end{table}

\textbf{2) Adversarial training with simple data augmentation could also yield effective robust models.} \Cref{fig:singular values (bcd)} shows that the singular value distribution of the perturbed graphs exhibits a right-shift. In adversarial training, adversarial examples can be randomly generated with the one having the largest GSE is selected as the worst-case perturbed graph for further training.
This is referred to as \textit{RndGSE}, a data augmentation method for adversarial training. According to \Cref{table:AT-GSE_cora} and \Cref{table:AT-GSE_citeseer}, i.e., the row where the model corresponds to RndGSE, we observe that adversarial training with RndGSE can yield robust GNN models with comparable adversarial accuracy performance, although it is inferior to AT-GSE. This is evident that GSE can be an effective indicator of GNN robustness.
\begin{table*}[!ht]
\renewcommand\arraystretch{1.2} 
\caption{Comparison between AT-GSE and baselines on Cora. The first column (Model) and the second column (AT, × means models without AT) show the names of the baselines and our proposed methods (light gray background) together. GPRGNN in the last line represents the backbone GNN model for AT-GSE. The second line is the attack ratio. The best result is highlighted in \textbf{bold}.}
\centering
\scalebox{0.92}{
\resizebox{\textwidth}{32mm}{
\begin{tabular}{c|c|c|c:c|c:c|c:c}
    \hline
    Model & AT & Clean & LRBCD & PRBCD & LRBCD & PRBCD & LRBCD & PRBCD  \\
    \hline
    & & 0\% & \multicolumn{2}{c|}{5\%} & \multicolumn{2}{c|}{10\%} & \multicolumn{2}{c}{25\%}  \\
    \hline
    GCN & × & 77.91±0.17 & 74.91±1.13 & 70.48±0.57 & 68.83±0.92 & 66.01±0.83 & 59.85±0.96 & 56.12±1.26 \\
    APPNP & × & 79.05±1.06 & 74.95±0.72 & 72.34±1.66 & 69.63±0.62 & 69.27±0.68 & 63.66±0.78 & 60.00±0.86 \\
    GPRGNN & × & 80.40±2.11 & 77.88±1.91 & 74.80±1.14 & 74.73±1.21 & 71.50±1.55 & 65.86±1.11 & 65.60±0.62 \\
    \hline
    & LRBCD  & 77.73±0.27 & 73.92±1.12 & 71.17±0.52 & 69.60±0.94 & 67.47±0.91 & 62.31±1.09 & 59.93±0.72 \\
    \multirow{-2}{*}{GCN} & PRBCD  & 80.15±0.15 & 75.42±0.96 & 73.59±0.45 & 69.85±0.79 & 69.41±0.44 & 62.31±0.95 & 60.48±0.64 \\
    & LRBCD  & 80.81±0.29 & 76.48±0.71 & 74.95±0.41 & 71.54±1.04 & 71.14±0.67 & 64.07±0.79 & 63.74±1.06 \\
    \multirow{-2}{*}{APPNP}   & PRBCD  & 81.72±0.38 & 76.59±1.08 & 73.77±0.52 & 70.88±1.61 & 69.38±0.64 & 62.09±0.62 & 63.52±0.74 \\
    & LRBCD  & 81.32±0.57 & 78.72±0.83 & 76.96±1.94 & 72.89±0.46 & 74.62±1.51 & 69.56±1.45 & 67.55±0.87 \\
    \multirow{-2}{*}{GPRGNN} & PRBCD  & 80.99±0.50 & 80.07±1.70 & 77.03±0.77 & 78.10±1.18 & 73.92±0.56 & 67.69±1.41 & 63.08±2.69 \\
    \hline
    \rowcolor{lightgray!30}
    GPRGNN & RndGSE & 80.84±0.33 & 79.82±0.20 & 77.47±0.38 & 77.51±0.60 & 74.58±0.66 & 73.30±0.60 & 70.33±0.66  \\ 
    \hline
    \rowcolor{lightgray!30}
    & GSE & 83.41±0.43 & 81.90±1.06 & 78.72±0.60 & 78.86±1.01 & 77.00±0.36 & 73.41±0.90 & 72.45±1.58 \\
    \cline{2-9}
    \rowcolor{lightgray!30}
    & RndSVD & \textbf{85.42±0.22} & \textbf{83.92±0.35} & 79.01±0.61 & \textbf{82.34±0.27} & 75.57±0.43 & \textbf{73.59±0.60} & 73.48±0.52 \\
    \cline{2-9}
    \rowcolor{lightgray!30}
    \multirow{-3}{*}{GPRGNN} & Nystr\"om & 84.69±0.59 & 81.43±0.40 & \textbf{81.39±0.39} & 79.89±0.60 & \textbf{80.00±0.52} & 73.15±1.57 & \textbf{76.26±1.82} \\
    \hline
\end{tabular}}}
\label{table:AT-GSE_cora}
\end{table*}

\begin{table*}[!ht]
\renewcommand\arraystretch{1.2} 
\caption{Comparison between AT-GSE and baselines on Citeseer. The first column (Model) and the second column (AT, × means models without AT) show the names of the baselines and our proposed methods (light gray background) together. GPRGNN in the last line represents the backbone GNN model for AT-GSE. The second line is the rate of budget. The best result is highlighted in \textbf{bold}.}
\centering
\scalebox{0.92}{
\resizebox{\textwidth}{32mm}{
\begin{tabular}{c|c|c|c:c|c:c|c:c}
    \hline
    Model & AT & Clean & LRBCD & PRBCD & LRBCD & PRBCD  & LRBCD & PRBCD  \\
    \hline
    &   & 0\% & \multicolumn{2}{c|}{5\%} & \multicolumn{2}{c|}{10\%} & \multicolumn{2}{c}{25\%} \\
    \hline
    GCN & × & 71.73±0.48 & 66.40±1.03 & 64.25±1.05 & 63.46±0.78 & 59.07±1.07 & 53.74±1.35 & 48.18±0.74 \\
    APPNP & × & 70.98±1.30 & 65.61±0.92 & 64.91±1.09 & 64.25±0.73 & 62.80±1.24 & 56.87±1.98 & 56.26±1.82 \\
    GPRGNN & × & 74.44±0.30 & 65.33±0.69 & 69.86±0.76 & 63.97±0.57 & 66.21±0.81 & 58.13±0.94 & 61.21±0.78 \\
    \hline
    \multirow{2}{*}{GCN} & LRBCD & 74.35±0.44 & 73.22±1.20 & 69.58±0.61 & 71.17±0.91 & 65.89±0.89 & 67.57±0.76 & 59.53±0.89 \\
    & PRBCD & 74.30±0.72 & 71.36±0.63 & 70.33±0.84 & 69.53±0.83 & 65.37±1.26 & 64.91±1.01 & 55.84±1.70 \\
    \multirow{2}{*}{APPNP} & LRBCD & 73.83±0.00 & 72.38±0.33 & 70.93±0.28 & 69.72±0.35 & 68.46±0.31 & 65.89±0.63 & 64.77±1.22 \\
    & PRBCD & 73.79±0.33 & 71.73±0.31 & 69.07±0.54 & 68.74±0.64 & 65.61±0.48 & 63.18±1.27 & 59.53±0.98 \\
    \multirow{2}{*}{GPRGNN} & LRBCD & 74.35±1.54 & 72.99±1.04 & 71.07±1.31 & 68.79±0.78 & 67.76±0.78 & 65.42±0.84 & 62.90±0.67 \\
    & PRBCD & 73.55±0.37 & 72.15±0.37 & 68.97±0.31 & 68.50±0.43 & 66.07±0.76 & 65.98±0.54 & 60.93±0.76 \\
    \hline
    \rowcolor{lightgray!30}
    GPRGNN & RndGSE & 75.14±0.46 & 68.32±0.19 & 73.74±0.65 & 67.85±0.50 & 72.71±0.37 & 65.09±0.55 & 71.59±0.50 \\
    \hline
    \rowcolor{lightgray!30}
    & GSE & 76.54±0.75 & \textbf{74.77±0.00} & 73.69±0.75 & \textbf{74.49±0.76} & 72.80±0.65 & \textbf{70.33±0.48} & 68.13±1.12 \\
    \cline{2-9}
    \rowcolor{lightgray!30} 
    & RndSVD & \textbf{77.62±0.74} & 73.41±0.44 & 73.79±0.68 & 72.15±0.81 & 73.41±0.85 & 69.67±0.25 & 71.07±0.85 \\
    \cline{2-9}
    \rowcolor{lightgray!30} 
    \multirow{-3}{*}{GPRGNN} & Nystr\"om & 76.68±0.14 & 74.02±0.56 & \textbf{74.95±0.31} & 71.92±0.57 & \textbf{74.11±0.63} & 68.60±0.65 & \textbf{73.08±1.03}  \\
    \hline
\end{tabular}}}
\label{table:AT-GSE_citeseer}
\end{table*}

\textbf{3) AT-GSE is effective against various adversarial attacks.} After random GSE (RndGSE) showed its excellent robustness, we refined the range of hyper-parameters in GSE and proposed the AT-GSE model. \Cref{table:AT_attacks} summarizes the robustness performance of AT-GSE when facing attacks on LRBCD, PRBCD, or PGD with APPNP and GPRGNN as the backbone models. As can be seen from the table, AT-GSE shows excellent robustness under various attacks, that is, it helps to resist a variety of adversarial attacks.

\textbf{4) AT-GSE is an effective defense method to improve GNN adversarial robustness.} The essential difference between AT-GSE and the existing ones is the way to generate adversarially perturbed graphs for inductive learning. 
\textit{AT-GSE as a gradient-based} method attempts to generate perturbations by adjusting the singular value distribution of the adjacency matrix of the clean graph, so that the adversarial training on such perturbed graph could track the structural properties of the underlying graph. As is evident from the results in \Cref{table:AT-GSE_cora}, \Cref{table:AT-GSE_citeseer}, \Cref{table:AT-GSE_cora_ml}, and \Cref{table:AT-GSE_SBM_homo}, we compare the empirical robustness of our proposed model with GCN, APPNP, and GPRGNN with and without adversarial training on Cora, Citeseer, Cora\_ML and SBM (homo.). The best accuracy is shown in bold in the tables, which clearly shows that our proposed AT-GSE outperforms other baselines in terms of robustness. In these experimental results, we observed that our proposed \textit{AT-GSE method outperforms other baselines in most cases}. All these tables show that under both LRBCD and PRBCD attacks, AT-GSE significantly improves the robustness of standard GNNs. Given that this method maintains the best accuracy even on clean data sets, i.e., unperturbed graphs, it is reasonable to verify that AT-GSE greatly improves the robustness of GNN while still maintaining the accuracy of classification.

\textbf{5) AT-GSE is selective with different GNN models to maintain its clean and adversarial accuracy on various datasets.} Our proposed AT-GSE algorithm works together with different GNN models, where different models could yield distinct performance.
The previous columns of AT-GSE in \Cref{table:AT-GSE_cora_ml} and \Cref{table:AT-GSE_SBM_homo} are marked with the benchmark GNN models, upon which the robust models are adversarially trained. Specifically, for the Cora\_ML dataset, GPRGNN and APPNP models with AT-GSE are more suitable to defend against PRBCD and LRBCD attacks, respectively, and for SBM (homo.), GCN with AT-GSE is the best choice.
\begin{table*}[!ht]
\renewcommand\arraystretch{1.2} 
\caption{Comparison between AT-GSE and baselines on Cora\_ML. The first column (Model) and the second column (AT, × means models without AT) show the names of the baselines and our proposed methods (light gray background) together. GPRGNN and APPNP in the last two lines represent the GNN models; these rows are GPRGNN and APPNP, respectively. The second line is the rate of budget. The best result is highlighted in \textbf{bold}.}
\centering
\scalebox{0.92}{
\resizebox{\textwidth}{34mm}{
\begin{tabular}{c|c|c|c:c|c:c|c:c}
    \hline
    Model & AT & Clean & LRBCD & PRBCD & LRBCD & PRBCD & LRBCD & PRBCD \\
    \hline
    & & 0\% & \multicolumn{2}{c|}{5\%} & \multicolumn{2}{c|}{10\%} & \multicolumn{2}{c}{25\%} \\
    \hline
    GCN & × & 84.12±0.11 & 77.15±0.58 & 74.19±0.67 & 73.20±1.01 & 68.24±0.58 & 67.46±1.21 & 57.04±2.28 \\
    APPNP & × & 82.29±0.23 & 80.67±1.11 & 74.33±0.79 & 77.85±1.00 & 67.89±1.33 & 73.10±1.10 & 56.90±1.68 \\
    GPRGNN & × & 81.44±0.32 & 75.14±0.92 & 71.55±0.90 & 70.74±0.43 & 67.50±2.08 & 63.66±1.02 & 60.35±2.72 \\
    \hline
    \multirow{2}{*}{GCN} & LRBCD & 85.56±0.00 & 80.67±0.53 & 75.81±0.67 & 78.06±0.91 & 69.96±1.11 & 71.41±0.83 & 56.16±0.48 \\
    & PRBCD & 85.49±0.21 & 79.30±0.56 & 76.51±0.52 & 76.23±0.96 & 70.00±1.02 & 70.46±1.26 & 57.43±0.91 \\
    \multirow{2}{*}{APPNP} & LRBCD & 86.62±0.35 & 80.67±0.69 & 77.89±0.74 & 78.27±0.88 & 72.68±0.82 & 74.58±0.90 & 62.61±0.70 \\
    & PRBCD & 85.11±1.32 & 79.65±1.54 & 78.94±0.44 & 77.89±1.23 & 73.10±0.87 & 73.38±1.32 & 61.37±0.61 \\
    \multirow{2}{*}{GPRGNN} & LRBCD & 81.37±0.11 & 78.31±0.82 & 76.13±0.85 & 75.53±0.48 & 71.83±0.59 & 68.87±0.85 & 63.66±0.87 \\
    & PRBCD & 81.02±0.56 & 77.15±1.93 & 76.13±0.34 & 73.45±1.26 & 71.41±0.26 & 67.61±1.32 & 62.43±0.52 \\
    \hline
    \rowcolor{lightgray!30}
    GPRGNN &  & 82.50±0.35 & 78.38±0.87 & 76.48±1.07 & 74.89±0.86 & 73.80±1.05 & 69.33±1.59 & 69.19±0.91 \\
    \cline{0-0} \cline{3-9}
    \rowcolor{lightgray!30}
    APPNP & \multirow{-2}{*}{GSE} & \textbf{87.75±0.14} & 81.37±0.68 & 78.42±0.45 & 78.94±0.70 & 72.71±0.50 & 74.89±0.83 & 60.67±0.89 \\
    \cline{0-0}\cline{2-9}
    \rowcolor{lightgray!30}
    GPRGNN &  & 86.16±0.55 & 82.61±0.59 & 78.31±0.42 & 79.05±0.84 & 77.25±0.55 & 70.07±1.55 & 68.77±1.40 \\
    \cline{0-0} \cline{3-9}
    \rowcolor{lightgray!30}
    APPNP & \multirow{-2}{*}{RndSVD} & 86.90±0.14 & \textbf{84.58±0.92} & 76.87±0.69 & \textbf{80.67±0.79} & 72.96±1.14 & \textbf{75.04±0.96} & 60.42±0.85 \\
    \cline{0-0}\cline{2-9}
    \rowcolor{lightgray!30}
    GPRGNN & Nystr\"om & 85.25±1.25 & 82.39±0.76 & \textbf{79.54±0.48} & 78.27±0.86 & \textbf{77.46±0.42} & 72.82±1.59 & \textbf{73.80±0.53} \\
    \hline
\end{tabular}}}
\label{table:AT-GSE_cora_ml}
\end{table*}

\begin{table*}[!ht]
\renewcommand\arraystretch{1.2} 
\caption{Comparison between AT-GSE and baselines on SBM(homo.). The first column (Model) and the second column (AT, × means models without AT) show the names of the baselines and our proposed methods (light gray background) together. GCN in the last line represents the backbone GNN model for AT-GSE. The second line is the rate of budget. The best model is highlighted in \textbf{bold}.}
\centering
\scalebox{0.92}{
\resizebox{\textwidth}{26mm}{
\begin{tabular}{c|c|c|c:c|c:c|c:c}
 \hline
    Model & AT & Clean & LRBCD & PRBCD & LRBCD & PRBCD & LRBCD & PRBCD \\
    \hline
    & & 0\% & \multicolumn{2}{c|}{5\%} & \multicolumn{2}{c|}{10\%} & \multicolumn{2}{c}{25\%} \\
    \hline
    GCN & × & 86.46±0.49 & 80.20±0.49 & 80.61±0.76 & 77.47±0.91 & 78.48±1.57 & 63.74±0.71 & 70.81±1.31 \\
    APPNP & × & 78.18±4.02 & 86.06±2.11 & 73.43±4.19 & 78.08±1.11 & 71.21±4.15 & 61.92±0.91 & 67.88±3.31 \\
    GPRGNN & × & 65.15±8.81 & 52.12±2.60 & 63.84±8.04 & 51.92±2.13 & 63.43±7.98 & 51.11±1.03 & 62.53±7.52 \\
    \hline
    & LRBCD & 86.77±0.30 & 81.72±0.54 & 84.14±0.65 & 79.39±0.49 & 82.83±1.20 & 68.59±0.84 & 80.30±0.51 \\
    \multirow{-2}{*}{GCN} & PRBCD & 86.87±0.00 & 84.55±0.65 & 85.35±0.51 & 81.11±0.46 & 84.34±0.81 & 71.62±0.71 & 82.12±0.91 \\
    & LRBCD & 86.26±3.70 & 81.72±1.15 & 79.60±2.38 & 76.06±1.57 & 76.67±2.84 & 62.83±0.99 & 70.71±1.69 \\
    \multirow{-2}{*}{APPNP} & PRBCD & 84.95±4.57 & 81.31±0.68 & 76.97±3.06 & 73.33±0.67 & 72.22±2.08 & 64.65±0.90 & 65.66±1.43 \\
    & LRBCD & 82.53±0.46 & 80.81±0.00 & 78.48±2.17 & 78.08±0.46 & 76.97±2.01 & \textbf{72.53±0.88} & 73.94±2.20 \\
    \multirow{-2}{*}{GPRGNN} & PRBCD & 83.03±5.66 & 79.60±0.40 & 77.07±4.50 & 73.74±0.78 & 75.05±4.24 & 63.84±0.88 & 70.30±4.89 \\
    \hline
    \rowcolor{lightgray!30}
    GCN & AT-GSE & \textbf{88.89±0.00} & \textbf{84.95±0.30} & \textbf{86.06±0.61} & \textbf{81.72±0.30} & \textbf{85.25±0.67} & 70.00±1.02 & \textbf{83.13±0.65} \\
    \hline
\end{tabular}}}
\label{table:AT-GSE_SBM_homo}
\end{table*}

\textbf{6) AT-GSE can also achieve good prediction results on heterophily graphs.} \Cref{table:AT-GSE_SBM_hetero} presents the performance of AT-GSE and baselines on a heterophily graph. It appears our proposed AT-GSE method significantly improves the robustness against adversarial attacks and the accuracy of clean datasets with a large margin. 
Remarkably, the adversarially trained GPRGNN model with AT-GSE achieves state-of-the-art performance on the clean SBM dataset.
\begin{table*}[!ht]
\renewcommand\arraystretch{1.2} 
\caption{Comparison between AT-GSE and baselines on SBM(hetero.). The first column (Model) and the second column (AT, × means models without AT) show the names of the baselines and our proposed methods (light gray background) together. GPRGNN in the last line represents the backbone GNN model for AT-GSE. The second line is the rate of budget. The best result is highlighted in \textbf{bold}.}
\centering
\scalebox{0.92}{
\resizebox{\textwidth}{26mm}{
\begin{tabular}{c|c|c|c:c|c:c|c:c}
    \hline
    Model & AT & Clean & LRBCD & PRBCD & LRBCD & PRBCD & LRBCD & PRBCD  \\
    \hline
    &   & 0\% & \multicolumn{2}{c|}{5\%} & \multicolumn{2}{c|}{10\%} & \multicolumn{2}{c}{25\%}  \\
    \hline
    GCN & × & 62.63±1.01 & 60.61±1.01 & 60.30±0.65 & 58.69±1.05 & 60.10±0.81 & 53.54±1.69 & 56.97±1.70  \\
    APPNP & × & 57.88±3.35 & 50.30±0.76 & 53.74±2.01 & 49.60±0.54 & 50.61±1.46 & 49.39±0.54 & 46.16±1.57  \\
    GPRGNN & × & 61.52±8.59 & 59.60±11.43 & 61.01±8.05 & 59.26±11.03 & 61.01±8.05 & 56.87±8.02 & 59.70±6.71 \\
    \hline
    & LRBCD  & 61.62±0.00 & 59.19±0.49 & 60.40±0.88 & 57.98±0.49 & 59.19±1.37 & 53.64±0.30 & 56.16±1.64 \\
    \multirow{-2}{*}{GCN} & PRBCD  & 64.65±0.00 & 61.52±0.54 & 64.04±0.81 & 59.70±0.84 & 63.03±1.12 & 52.73±1.74 & 61.01±0.81 \\
    & LRBCD & 49.49±0.00 & 49.49±0.00 & 49.49±0.00 & 49.49±0.00 & 49.49±0.00 & 49.49±0.00 & 49.49±0.00  \\
    \multirow{-2}{*}{APPNP} & PRBCD  & 64.04±5.44 & 59.39±2.29 & 49.49±0.00 & 55.66±0.95 & 49.49±0.00 & 43.54±1.23 & 49.49±0.00 \\
    & LRBCD  & 82.73±1.39 & 82.63±1.34 & 79.49±1.87 & 75.96±0.76 & 71.72±5.21 & 65.66±3.94 & 68.79±4.05  \\
    \multirow{-2}{*}{GPRGNN} & PRBCD  & 86.26±1.98 & 84.55±0.65 & 81.72±1.78 & 75.15±0.49 & 79.39±1.64 & 62.12±2.13 & 74.44±1.75   \\
    \hline
    \rowcolor{lightgray!30}
    GPRGNN & AT-GSE & \textbf{90.40±1.37} & \textbf{87.88±0.64} & \textbf{84.75±1.78} & \textbf{79.49±1.28} & \textbf{81.21±2.72} & \textbf{66.36±4.29} & \textbf{75.86±1.89} \\ 
    \hline
\end{tabular}}}
\label{table:AT-GSE_SBM_hetero}
\end{table*}

\textbf{7) GNN models with AT-GSE can not only improve adversarial robustness but also help generalization ability.} 
\Cref{table:AT-GSE_cora} shows that our proposed AT-GSE method still has stronger prediction performance on clean graphs, even though the models are adversarially trained. It can be concluded that AT-GSE not only improves adversarial accuracy but also enhances clean accuracy. To further certify the generalization ability of the AT-GSE model, we specifically tested the accuracy of backbone models (GCN, APPNP, and GPRGNN) with and without AT-GSE on different 5 clean graphs, as shown in \Cref{table:AT-GSE_clean_graphs}. It can be seen that the performance of AT-GSE based on the three models of GCN, APPNP, and GPRGNN is better than the original normally-trained ones. As such, we conclude that AT-GSE has strong generalization ability as it can be employed by any backbone model, and its superiority to the original model shows that it can improve the classification accuracy of the model on clean graphs.

To sum up,
based on the experimental results of the naturally and adversarially trained GNN models on various graph datasets, we conclude that AT-GSE consistently improves adversarial robustness and clean accuracy not only on homophilic datasets but also on heterophilic graphs. Therefore, AT-GSE can be widely employed to train robust and generalizable GNN models to mine different types of graph-structured data.

\begin{table*}[!ht]
\renewcommand\arraystretch{1.2} 
\caption{The original model and AT-GSE on clean graphs.}
\centering
\scalebox{0.96}{
\begin{tabular}{c|c|c|c|c|c|c}
\hline
    Model & AT & Cora & Cora\_ML & Citeseer & SBM(homo.) & SBM(hetero.) \\ \hline
    \multirow{2}{*}{GCN} & × & 77.91±0.17 & 84.12±0.11 & 71.73±0.48 & 86.46±0.49 & 62.63±1.01 \\ \cline{2-7}
    & AT-GSE & \textbf{77.95±0.61} & \textbf{86.13±0.92} & \textbf{74.35±0.74} & \textbf{88.89±0.00} & \textbf{64.85±1.48} \\ \hline
    \multirow{2}{*}{APPNP} & × & 79.05±1.06 & 85.53±0.87 & 70.98±1.30 & 78.18±4.02 & 57.88±3.35 \\ \cline{2-7} 
    & AT-GSE & \textbf{82.64±0.18} & \textbf{87.75±0.14} & \textbf{73.13±0.60} & \textbf{80.51±4.50} & \textbf{60.61±2.78} \\ \hline
    \multirow{2}{*}{GPRGNN} & × & 80.40±2.11 & 81.44±0.32 & 63.55±1.40 & 65.15±8.81 & 61.52±8.59 \\ \cline{2-7} 
    & AT-GSE & \textbf{83.41±0.43} & \textbf{83.27±0.36} & \textbf{76.54±0.75} & \textbf{86.57±2.22} & \textbf{90.40±1.37} \\ \hline
\end{tabular}}
\label{table:AT-GSE_clean_graphs}
\end{table*}

\subsection{Scalability Performance}
Speaking of scalability performance, it is necessary to consider aspects such as \textit{computing efficiency, data size, storage overhead, and prediction accuracy}.

Based on the above experiments, we can conclude that AT-GSE has excellent robustness and generalization and is applicable to various types of graph data. However, existing studies reveal that adversarial training is difficult to be applied to large-scale graphs. What's worse is that AT-GSE deals with singular value distribution, which may further impose the computational burden and prevent it from the potential applications to larger graph datasets. To resolve this issue, we propose two variants (RndSVD and Nystr\"om) of AT-GSE models to improve the model's training and classifying capabilities on larger-scale graph data.

As described in \Cref{sub:RndSVD}, RndSVD leverages random projection of the original large-sized adjacency matrix onto a low-dimensional subspace so as to reduce the computational complexity of SVD operations. The Nystr\"om method is to calculate a low-rank approximation of the original adjacency matrix, avoiding the need of SVD operations (see \Cref{sub:Nystrom} for details). Both methods have theoretical guarantee of the approximation accuracy with substantially reduced computational complexity. 

To demonstrate their practical usefulness, we randomly generated an adjacency matrix of a graph with 1,000 nodes to calculate the running time taken by the original SVD, RndSVD, and Nystr\"om for every epoch. Experimental results show that SVD takes 3.9 seconds per epoch, RndSVD needs 0.1 seconds, and Nystr\"om only consumes 0.05 seconds. It appears that \textbf{the running time of RndSVD and Nystr\"om methods are nearly 40 times and 80 times faster than the vanilla SVD respectively.}
In addition, we directly select the fastest method to test on the Pubmed dataset with about 20,000 nodes, which will be out-of-memory when running the original AT-GSE on a 40G GPU. The test outcomes are recorded in \Cref{table:GSE_pubmed}. Also using the 40G GPU, Nystr\"om can train the classify prediction results, most of which are better than our baselines. It is evident that \textbf{our proposed variants take up less storage space and also have great robustness on large-scale datasets.} 
\begin{table*}[!ht]
\renewcommand\arraystretch{1.2} 
\caption{Comparison between AT-GSE with Nystr\"om and baselines on Pubmed. The first column (Model) and the second column (AT, × means models without AT) show the names of the baselines and our proposed methods (light gray background) together. GPRGNN in the last line represents the backbone GNN model for AT-GSE. The second line is the attack ratio. The best result is highlighted in \textbf{bold}.}
\centering
\scalebox{0.92}{
\resizebox{\textwidth}{26mm}{
\begin{tabular}{c|c|c|c:c|c:c|c:c}
    \hline
    Model & AT & Clean & LRBCD & PRBCD & LRBCD & PRBCD & LRBCD & PRBCD \\
    \hline
    & & 0\% & \multicolumn{2}{c}{5\%} & \multicolumn{2}{c}{10\%} & \multicolumn{2}{c}{25\%} \\ 
    \hline
    GCN & × & 72.45±0.06 & 67.29±0.15 & 69.19±0.25 & 65.54±0.24 & 61.56±0.16 & 62.00±0.34 & 47.30±0.60 \\
    APPNP & × & 74.45±0.10 & 69.35±0.23 & 71.91±0.17 & 67.71±0.26 & 65.72±0.29 & 63.61±0.33 & 53.61±0.42 \\
    GPRGNN & × & 74.50±0.90 & 70.72±0.90 & 73.60±0.48 & 68.16±0.73 & 69.94±0.90 & 63.64±0.98 & 64.71±1.54 \\
    \hline
    & LRBCD & 76.06±0.08 & 70.81±0.17 & 70.73±0.26 & 69.03±0.37 & 63.19±0.36 & 65.53±0.34 & 49.22±0.53 \\
    \multirow{-2}{*}{GCN} & PRBCD & 72.48±0.00 & 68.19±0.21 & 70.15±0.24 & 66.62±0.19 & 63.43±0.27 & 63.21±0.34 & 49.45±0.52 \\
    & LRBCD & 72.70±0.10 & 69.59±0.26 & 69.22±0.25 & 68.10±0.19 & 64.76±0.43 & 64.97±0.21 & 54.92±0.33 \\
    \multirow{-2}{*}{APPNP} & PRBCD & 77.54±0.39 & 72.87±0.22 & 73.45±0.34 & 70.28±0.42 & 68.46±0.27 & 66.15±0.41 & 57.97±0.36 \\
    & LRBCD & 75.62±0.27 & 72.73±0.35 & 71.32±0.34 & 71.48±0.25 & 68.10±0.26 & \textbf{69.32±0..16} & 62.46±0.50 \\
    \multirow{-2}{*}{GPRGNN} & PRBCD & 75.72±0.31 & 73.22±0.52 & 74.14±0.23 & 71.06±0.68 & 70.52±0.44 & 68.24±0.83 & 64.48±0.46 \\
    \hline
    \rowcolor{lightgray!30} 
    GPRGNN & Nystr\"om & \textbf{79.22±0.00} & \textbf{75.43±0.16} & \textbf{74.24±0.17} & \textbf{72.70±0.21} & \textbf{70.94±0.35} & 69.26±0.11 & \textbf{66.27±0..14} \\
    \hline
\end{tabular}}}
\label{table:GSE_pubmed}
\end{table*}

After determining the computational efficiency of AT-GSE with RndSVD and Nystr\"om, we need to further verify that they also have excellent robust performance. Therefore, we compared these two methods of AT-GSE with baselines on the Cora, Citeseer and Cora\_ML datasets too. The results are also shown in \Cref{table:AT-GSE_cora}, \Cref{table:AT-GSE_citeseer} and \Cref{table:AT-GSE_cora_ml}. These two variants substantially reduce time complexity at the cost of slightly degraded robustness performance. \textbf{It is worth noting that they have almost the same performances as the original AT-GSE.} Specifically, the prediction accuracy of AT-GSE using RndSVD or Nystr\"om is greater than that of baselines, indicating that they are end up with more robust GNN models. The strongest effect based on different backbone models shows that they are also selective to different backbone models. Moreover, the applicability to different datasets reflects good generalization ability. In other words, these two AT-GSE methods, especially Nystr\"om, yield excellent robustness, generalization, and scalability performance, and can be a promising candidate of robust GNN training methods.

\begin{figure*}[!ht]
    \centering
    \subfloat{\includegraphics[width=3.5in]{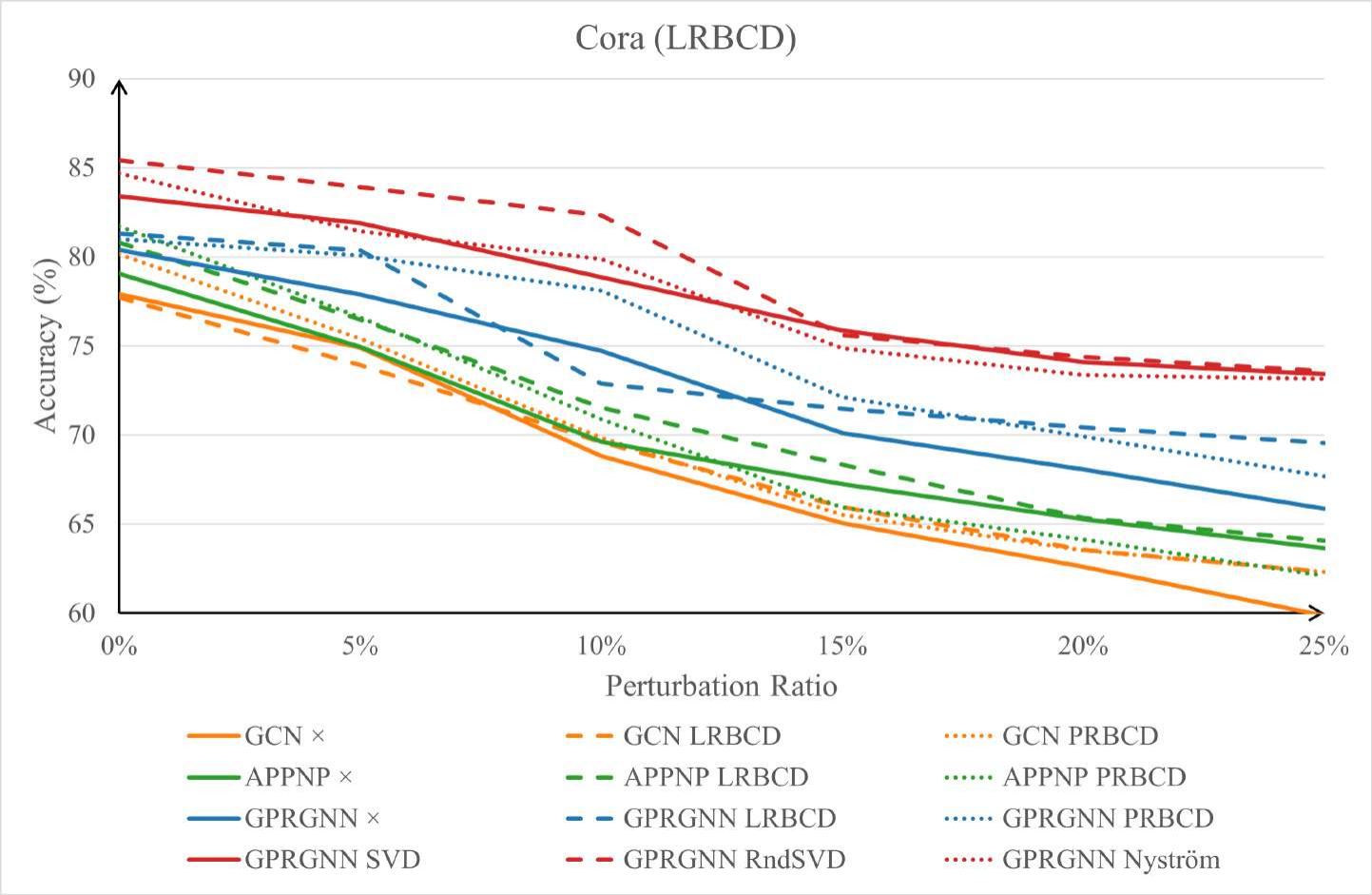}%
    \label{minifig:cora_LR}}
    \hfil
    \subfloat{\includegraphics[width=3.5in]{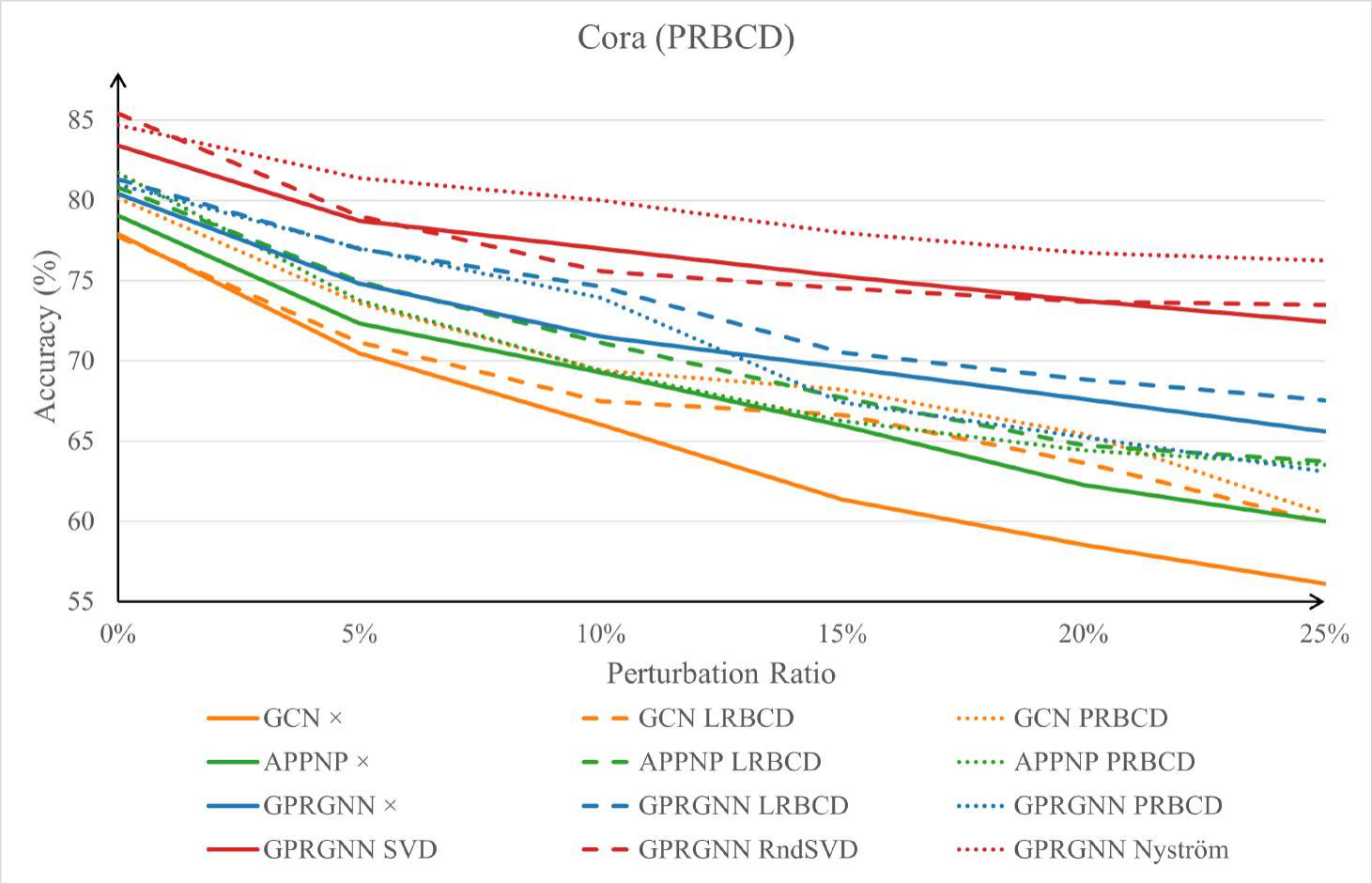}%
    \label{minifig:cora_PR}}
    \caption{The accuracy of each model on the Cora dataset under LRBCD (left) or PRBCD (right) attacks. In the legend, the former is the backbone model and the latter is the name of the method for generating adversarial examples. For example, GPRGNN RndSVD is an adversarial training with GPRGNN as the backbone model and RndSVD generating adversarial examples. The red line in the figure is the method we proposed, and the rest are baselines.}
    \label{fig:Cora_performance}
\end{figure*}

\begin{figure*}[!ht]
    \centering
    \subfloat{\includegraphics[width=3.5in]{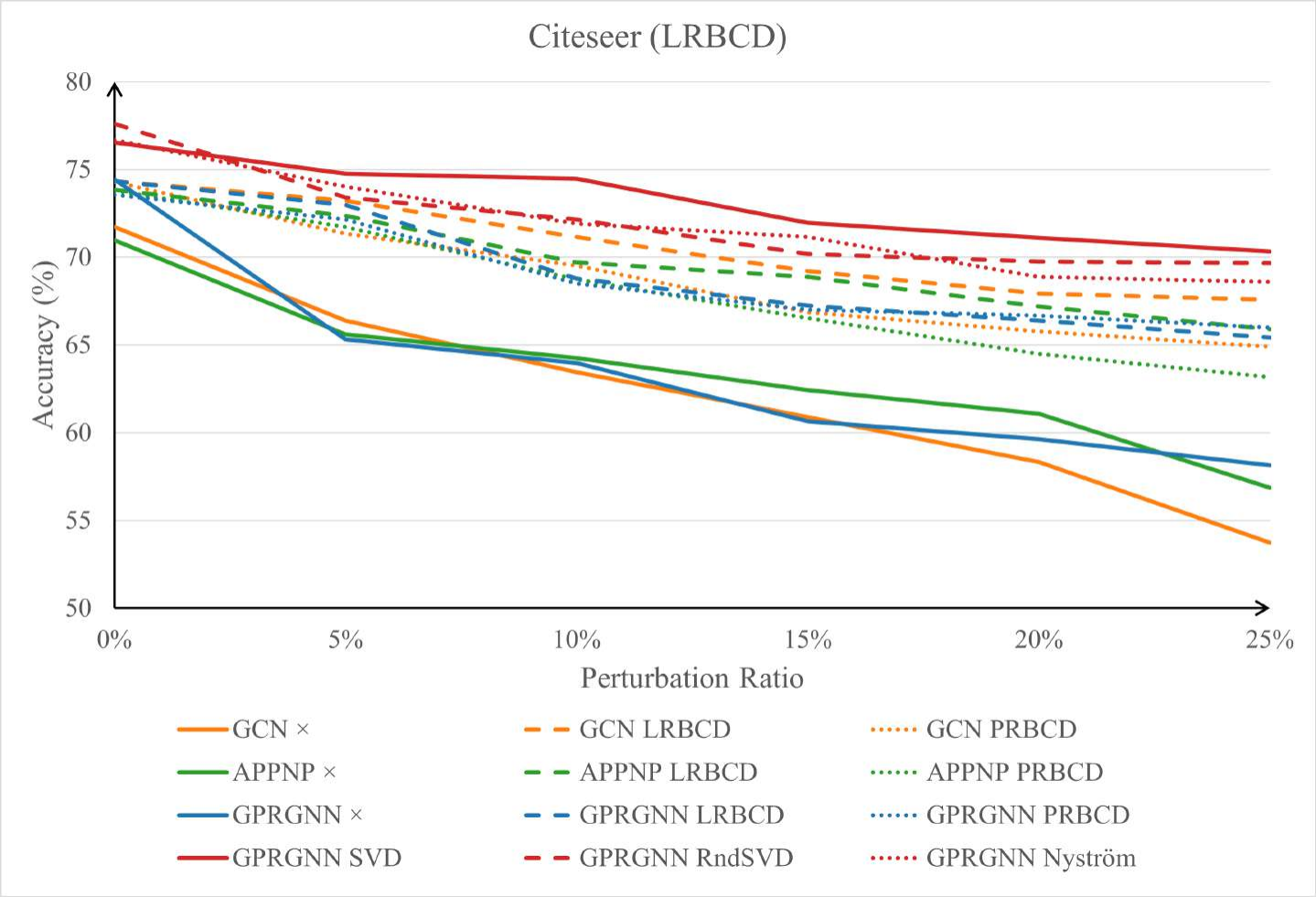}%
    \label{minifig:citeseer_LR}}
    \hfil
    \subfloat{\includegraphics[width=3.5in]{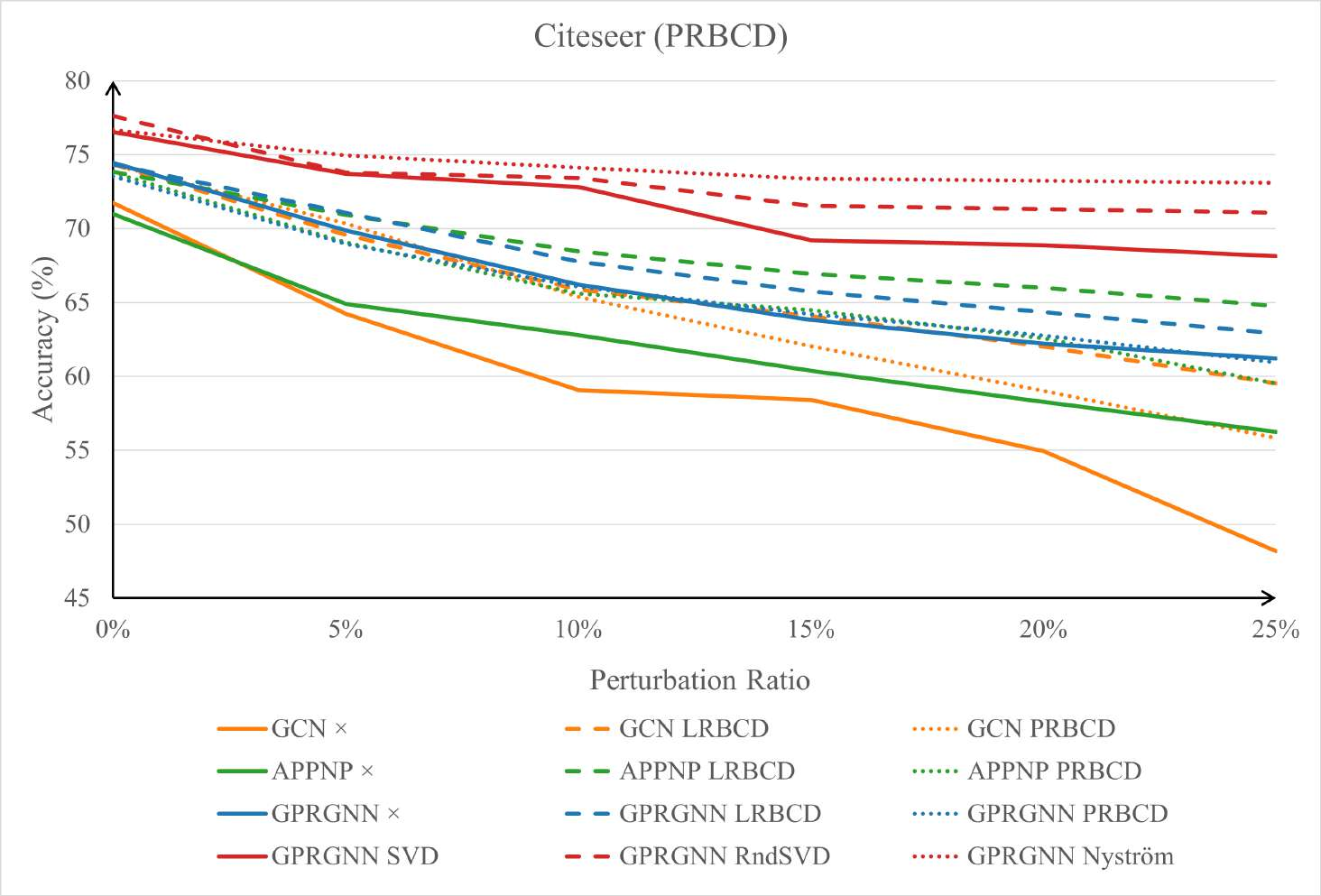}%
    \label{minifig:citeseer_PR}}
    \caption{The accuracy of each model on the Citeseer dataset under LRBCD (left) or PRBCD (right) attacks. In the legend, the former is the backbone model and the latter is the name of the method for generating adversarial examples. For example, GPRGNN RndSVD is an adversarial training with GPRGNN as the backbone model and RndSVD generating adversarial examples. The red line in the figure is the method we proposed, and the rest are baselines.}
    \label{fig:Citeseer_performance}
\end{figure*}

\begin{figure*}[!ht]
    \centering
    \subfloat{\includegraphics[width=3.5in]{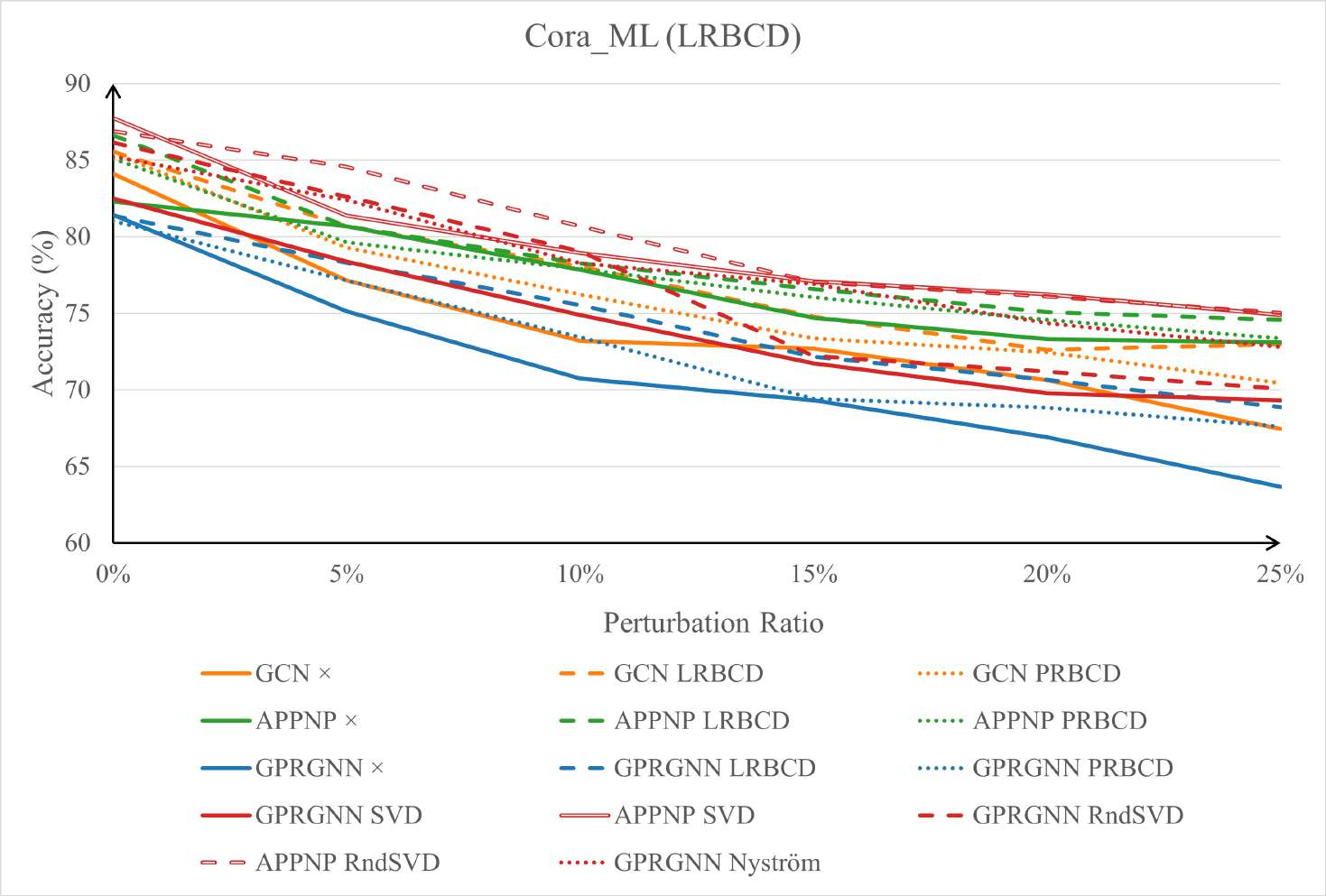}%
    \label{minifig:cora_ml_LR}}
    \hfil
    \subfloat{\includegraphics[width=3.5in]{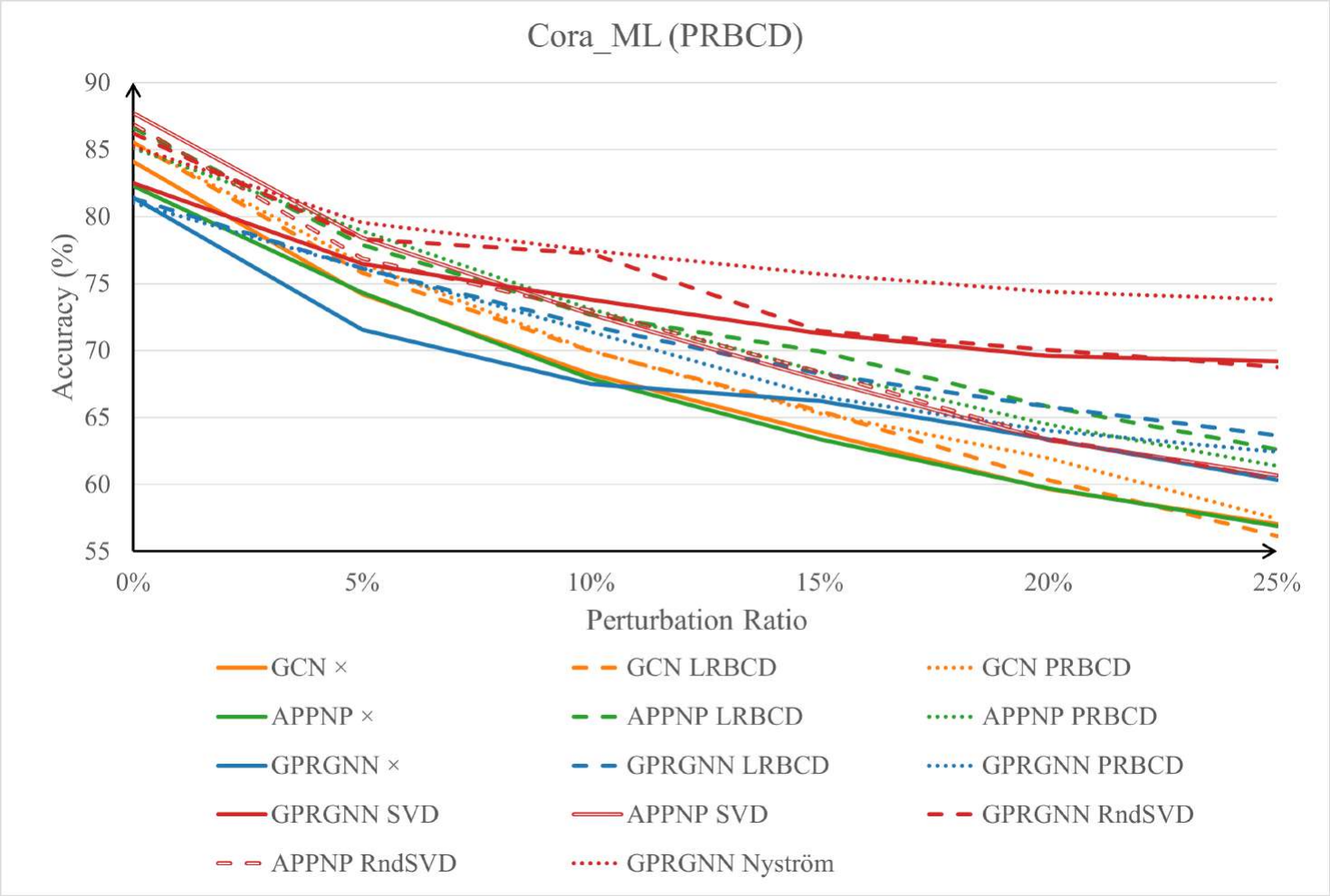}%
    \label{minifig:cora_ml_PR}}
    \caption{The accuracy of each model on the Cora dataset under LRBCD (left) or PRBCD (right) attacks. In the legend, the former is the backbone model and the latter is the name of the method for generating adversarial examples. For example, GPRGNN RndSVD is an adversarial training with GPRGNN as the backbone model and RndSVD generating adversarial examples. The red line in the figure is the method we proposed, and the rest are baselines.}
    \label{fig:Cora_ML_performance}
\end{figure*}

More interestingly, AT-GSE with RndSVD and Nystr\"om exhibits different behaviors in defending against local and global topology attacks. According to the comparison of these two methods, we can find that RndSVD is more dedicated to the local graph properties compared to Nystr\"om. It retains all graph properties related to the low-rank parts, discards unnecessary high-rank parts, and only adjusts the medium-parts that are most vulnerable to topology attacks. In contrast, the Nystr\"om method is of high simplicity without time-consuming SVD operations. It modifies the graph's global properties by adjusting the low-rank/low-frequency graph spectrum as a whole. Therefore, when they are applied to AT-GSE, \textbf{RndSVD is expected to defend local attacks, while Nystr\"om is more favorable global attacks.} Evidently, in \Cref{fig:Cora_performance}, \Cref{fig:Citeseer_performance}, and \Cref{fig:Cora_ML_performance}, for local LRBCD attacks, SVD or RndSVD is better, while Nystr\"om performs better against global PRBCD attacks.

Finally, we also tested the prediction results of AT-GSE with Nystr\"om on the larger heterophily graph, e.g., the WikiCS dataset. \Cref{table:GSE_wikics} shows that the AT-GSE-like method that reduces computational complexity also shows \textbf{its strong robust performance for heterophily graphs}, outperforming the existing state-of-the-art methods.

\begin{table*}[!ht]
\renewcommand\arraystretch{1.2} 
\caption{Comparison between AT-GSE with Nystr\"om and baselines on WikiCS. The first column (Model) and the second column (AT, × means models without AT) show the names of the baselines and our proposed methods (light gray background) together. GPRGNN in the last line represents the backbone GNN model for AT-GSE. The second line is the attack ratio. The best result is highlighted in \textbf{bold}.}
\centering
\scalebox{0.92}{
\resizebox{\textwidth}{26mm}{
\begin{tabular}{c|c|c|c:c|c:c|c:c}
    \hline
    Model & AT & Clean & LRBCD & PRBCD & LRBCD & PRBCD & LRBCD & PRBCD \\
    \hline
    & & 0\% & \multicolumn{2}{c}{5\%} & \multicolumn{2}{c}{10\%} & \multicolumn{2}{c}{25\%} \\ 
    \hline
    GCN & × & 73.13±0.00 & 50.75±0.47 & 51.07±0.34 & 44.86±0.50 & 45.29±0.43 & 39.85±0.35 & 36.85±0.26 \\
    APPNP & × & 73.00±1.11 & 59.06±1.45 & 56.72±0.91 & 55.00±1.84 & 51.31±0.73 & 48.33±1.41 & 43.44±1.23 \\
    GPRGNN & × & 72.69±0.00 & 60.33±0.41 & 62.17±0.64 & 57.03±0.24 & 58.41±0.34 & 53.07±0.39 & 53.14±0.50 \\
    \hline
    & LRBCD & 76.22±0.44 & 63.52±0.68 & 59.82±0.27 & 60.12±0.49 & 53.84±0.32 & 56.56±0.47 & 44.51±0.47 \\
    \multirow{-2}{*}{GCN} & PRBCD & 75.90±0.61 & 59.70±0.73 & 57.97±0.37 & 55.01±0.45 & 51.81±0.50 & 51.04±0.50 & 42.48±0.40 \\
    & LRBCD & 75.39±0.24 & 65.63±0.59 & 64.78±0.34 & 62.24±0.42 & 60.54±0.34 & 58.43±0.54 & 52.71±0.52 \\
    \multirow{-2}{*}{APPNP} & PRBCD & 75.99±0.41 & 63.77±0.56 & 62.29±0.79 & 60.63±0.74 & 57.54±0.50 & 56.09±0.89 & 50.33±0.51 \\
    & LRBCD & 73.92±0.51 & 64.43±0.67 & 67.42±0.51 & 63.14±0.91 & 66.05±0.54 & \textbf{61.91±0.67} & \textbf{63.36±0.60} \\
    \multirow{-2}{*}{GPRGNN} & PRBCD &76.04±0.67 & 66.84±0.73 & 68.00±0.77 & 64.22±1.15 & 65.88±0.82 & 61.36±1.20 & 61.91±0.83 \\
    \hline
    \rowcolor{lightgray!30} 
    GPRGNN & Nystr\"om & \textbf{78.64±0.63} & \textbf{67.58±1.19} & \textbf{69.36±0.49} & \textbf{64.66±1.10} & \textbf{66.68±0.39} & 61.46±0.28 & 61.54±0.95 \\
    \hline
\end{tabular}}}
\label{table:GSE_wikics}
\end{table*}

To conclude, by leveraging graph subspace energy as a regularizer in the adversarial training of GNN models on inductive learning, AT-GSE possesses excellent robustness, generalization, and scalability performance and should be of practical significance.

\section{Conclusion}
\label{sec:conclusion}
We proposed an adversarial training with graph subspace energy (AT-GSE) method to improve GNN robustness to adversarial perturbations in the node classification task on inductive learning. 
Our proposed AT-GSE method improves GNN adversarial robustness by protecting the singular value distribution of the adjacency matrix, which is vulnerable to adversarial topology attacks in the inference phase. It turns out that GSE is a reasonable measurable indicator of GNN robustness and a suited control of GSE could yield more robust GNN models.
We also proposed AT-GSE with RndSVD and Nystr\"om approximation to defend against the local and global topology perturbations respectively, with enhanced computational efficiency. 
Extensive experiments conducted on 7 datasets in an inductive learning setting demonstrate the superiority of our proposed AT-GSE method in robustness, generalization, and scalability to the state-of-the-art methods.
Theoretical underpinnings of GSE towards GNN adversarial robustness and the scalability to larger-sized graph datasets are interesting yet challenging open questions for future works.

\bibliographystyle{IEEEtran}


\end{document}